\documentclass[10pt,journal,compsoc]{IEEEtran}
\usepackage{changes}
\usepackage{color}
\usepackage{soul}
\soulregister\cite7
\soulregister\ref7 
\definechangesauthor[name={Author1}, color=red]{gy}

\ifCLASSOPTIONcompsoc
\usepackage[nocompress]{cite}
\else
\usepackage{cite}
\fi
\ifCLASSINFOpdf

\else

\fi

\usepackage{graphicx}
\usepackage{subfigure}
\usepackage{bm}
\usepackage{amsthm,amsmath,amssymb}
\usepackage{multirow}
\usepackage{ragged2e}
\newtheorem{theorem}{Theorem}
\newtheorem{corollary}[theorem]{Corollary}

\newtheorem{assumption}[theorem]{Assumption}
\newtheorem{definition}[theorem]{Definition}
\usepackage[ruled]{algorithm2e}
\allowdisplaybreaks[4]

\begin{document}
	
	\title{Complementary to Multiple Labels: A Correlation-Aware Correction Approach}
	
	\author{Yi~Gao,~
		Miao~Xu,
		and~Min-Ling~Zhang,~\IEEEmembership{Senior Member,~IEEE}
		\IEEEcompsocitemizethanks{\IEEEcompsocthanksitem Yi Gao is with the School of Cyber Science and Engineering, Southeast University, Nanjing 210096, China and the Key Laboratory of Computer Network and Information Integration (Southeast University), Ministry of Education, China. E-mail: gao\_yi@seu.edu.cn
			\IEEEcompsocthanksitem Miao Xu is with The University of Queensland, Australia. E-mail: miao.xu@uq.edu.au
			\IEEEcompsocthanksitem Min-Ling Zhang (corresponding author) is with the School of Computer Science and Engineering,Southeast University, Nanjing 210096, China and the Key Laboratory of Computer Network and Information Integration (Southeast University), Ministry of Education, China. E-mail: zhangml@seu.edu.cn
		}
		\thanks{Manuscript received April 19, 2005; revised August 26, 2015.}}
	
	%
	%

\markboth{Journal of \LaTeX\ Class Files,~Vol.~14, No.~8, August~2015}%
{Yi Gao \MakeLowercase{\textit{et al.}}: Complementary to Multiple Labels: A Correlation-Aware Correction Approach}
%


\IEEEtitleabstractindextext{%
	\begin{abstract}
		\textit{Complementary label learning} (CLL) requires annotators to give \emph{irrelevant} labels instead of relevant labels for instances. Currently, CLL has shown its promising performance on multi-class data by estimating a transition matrix. However, current multi-class CLL techniques cannot work well on multi-labeled data since they assume each instance is associated with one label while each multi-labeled instance is relevant to multiple labels. Here, we show theoretically how the estimated transition matrix in multi-class CLL could be distorted in multi-labeled cases as they ignore co-existing relevant labels. Moreover, theoretical findings reveal that calculating a transition matrix from label correlations in \textit{multi-labeled CLL} (ML-CLL) needs multi-labeled data, while this is unavailable for ML-CLL. To solve this issue, we propose a two-step method to estimate the transition matrix from candidate labels. Specifically, we first estimate an initial transition matrix by decomposing the multi-label problem into a series of binary classification problems, then the initial transition matrix is corrected by label correlations to enforce the addition of relationships among labels. We further show that the proposal is classifier-consistent, and additionally introduce an MSE-based regularizer to alleviate the tendency of BCE loss overfitting to noises. Experimental results have demonstrated the effectiveness of the proposed method.
	\end{abstract}
	
	}

\maketitle

\IEEEdisplaynontitleabstractindextext

\IEEEpeerreviewmaketitle

\IEEEraisesectionheading{\section{Introduction}
	\label{sec:introduction}}

\begin{figure}
	\centering
	\includegraphics[width=250px]{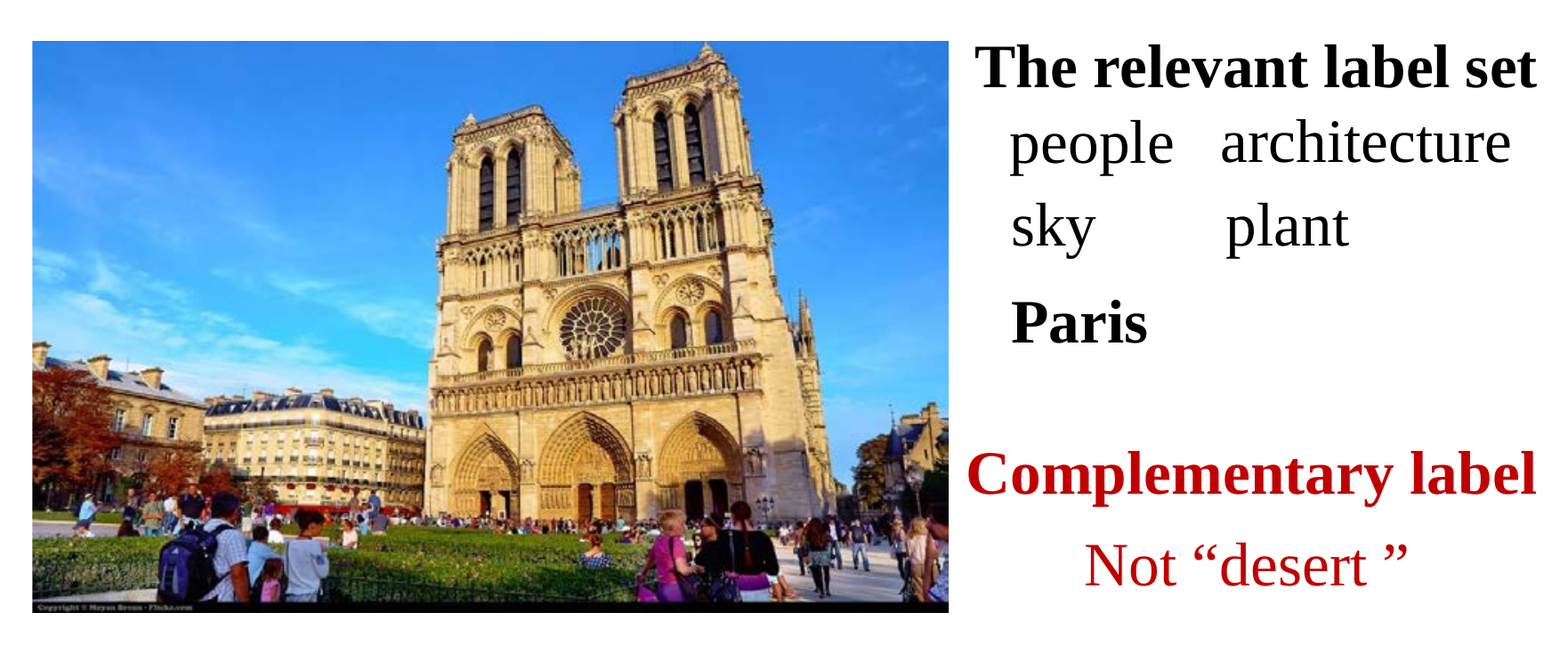}
	\vskip -0.15in
	\caption{An example of ML-CLL. The relevant labels of the image are \emph{people}, \emph{architecture}, \emph{sky}, \emph{plant}, and \textit{Paris}, while \emph{desert} is the complementary label of this image. The label \emph{Paris} is a complex semantic label, because it is difficult to be directly identified without domain knowledge.}
	\label{fig_1}
	\vskip -0.1in
\end{figure}

\noindent In \textit{multi-label learning} (MLL), each instance is associated with a set of relevant labels, where the learned classifier aims to predict all relevant labels of unseen instances \cite{ZhangZ14, lift_ZhangW15}. MLL is widely used in many real-world applications, such as text categorization \cite{ml/RubinCSS12, aaai/Tang0XPWC20}, image retrieval \cite{lambrecht2013does}, etc. However, collecting precisely multi-labeled data is laborious because of the unknown number of relevant labels per instance and the existence of complex semantic labels. For the example image in Fig. \ref{fig_1}, besides the label \textit{Architecture}, there exist other relevant labels whose accurate annotation needs one-by-one checking of the whole label space; in addition, annotators need special geographical and cultural domain knowledge to accurately label the image as \textit{Paris}.

To release the laborious of annotating multi-labeled data, we explore the problem setting of \textit{multi-labeled CLL} (ML-CLL), where each instance is associated with a \textit{single} complementary label (an irrelevant label of the instance) instead of multiple relevant labels. Providing such weakly supervised information will ease the labeling process in large label space because selecting one complementary label is low-cost and requires less domain knowledge than selecting all relevant labels. One example of ML-CLL is given in Fig. \ref{fig_1} when selecting \textit{desert} as the complementary label. Given the complementary label, the goal of ML-CLL is still the same as fully supervised MLL, i.e., learning a model that can accurately predict multiple relevant labels for unseen instances. 

The setting of CLL was initially applied in the multi-class learning task \cite{nips/IshidaNHS17, icml/IshidaNMS19,eccv/yu_learning_2018, icml/Chou0LS20,icml/GaoZ21,ijcai/WangFZ21,icml/FengK000S20}. Previous multi-class CLL approaches are based on an estimated transition matrix that summarizes the probability of a label being selected as a complementary label \cite{nips/IshidaNHS17,icml/IshidaNMS19,eccv/yu_learning_2018}. Although they have achieved a promising performance on multi-class data, they are restricted to the case where an instance is associated with only one relevant label. In this case, multi-class CLL approaches only consider the exclusive relationship among labels, while these approaches ignore that labels can bear other relationships in the multi-labeled case, especially the co-occurrence of labels. In fact,  relationships among labels are crucial to solving ML-CLL problems since the selection of a complementary label of an instance in MLL is the combined result against multiple relevant labels rather than against only a single relevant label. Misusing a technique targeting against a single relevant label to the multiple relevant labels case will result in a wrongly estimated transition matrix.

In this paper, we first theoretically analyze how the estimation of the transition matrix using the current multi-class CLL techniques could be distorted in multi-labeled cases. According to these findings, we observe that estimating the transition matrix in ML-CLL from label correlations needs to know relevant labels of instances, while these are unavailable. To remove this pain, we propose a two-step method to estimate the transition matrix in ML-CLL from candidate labels which are the complement of complementary labels. Our strategy includes: (1) estimating an initial transition matrix by decomposing the multi-label problem into binary classification problems; (2) using label correlations to correct the initial transition matrix by enforcing the addition of relationships among labels. The fast convergence of \textit{Cross-Entropy} (CE) loss benefits from focusing on instances that are difficult to classify, which may result in CE loss overfitting to noisy labeled data. As a type of CE loss, \textit{Binary CE} (BCE) loss has the same problem. The study of \cite{aaai/GhoshKS17} indicates that \textit{Mean square error} (MSE) loss is less sensitive to noisy labels than CE loss. As \textit{Binary CE} (BCE) loss is a benchmark of our approach, an MSE-based regularizer is further introduced to alleviate the tendency of it overfitting to noises. 

In addition, we show that our proposed ML-CLL can be easily combined with learning from relevant labels, which significantly extends the application scenario of the proposed algorithm. This combination is particularly useful, e.g. when labels are collected via crowdsourcing\cite{sindlinger2010crowdsourcing} where crowdworkers are asked to randomly select a complementary label and one or more relevant labels for an instance. Experimental results on various datasets demonstrate the effectiveness of the proposed approach. Especially in situation when each instance is only equipped with a complementary label and a relevant label, our proposal has superior performance, even comparable with the performance on fully supervised data. Our main contributions are summarized as follows:

\begin{itemize}
	\item We theoretically analyze the distortion of the transition matrix estimated by multi-class CLL in multi-labeled cases, because multi-class CLL techniques ignore the co-existence of relevant labels. Theoretical findings reveal that multi-labeled data is indispensable for calculating the transition matrix from label correlations.
	\item To solve the problem of unavailable multi-labeled data, we propose a two-step method to estimate the transition matrix from candidate labels. Moreover, we show theoretically that the proposed approach is classifier-consistent under a mild assumption.
	\item We introduce a practical strategy -- MSE-based regularization -- to alleviate the overfitting tendency of BCE loss. Our empirical study shows that the proposal obtains comparable performance with state-of-the-art baselines, which proves the effectiveness of our approach.
\end{itemize}

The rest of this paper are organized as follows. Section \ref{sec:related} briefly reviews related work of ML-CLL. Then we formalize the ML-CLL problem in Section \ref{sec:setup}, analyze it theoretically and describe our approach in Section \ref{sec:approach}. In Section \ref{sec:mse}, we introduce an MSE-based regularization and show how to adapt our method to bear an additional small amount of relevant labels. The experimental results are given in Section \ref{sec:experiments} and we conclude in Section \ref{sec:conclusion}. 

\section{Related Work}
\label{sec:related}

In this section, we will give a brief review of related work of ML-CLL, including MLL, \textit{partial multi-label learning} (PML) and multi-class CLL.

\subsection{Multi-Label Learning}
MLL problems aim to train a classifier that can predict a set of relevant labels for an unseen instance, where each training instance is associated with multiple relevant labels simultaneously. With the complexity of label correlation, the previous studies can be grouped into three categories\cite{tkde/ZhangZ14, tbd/WuWZYLZZ15, cvpr/BucakJJ11, jmlr/LiuTM17}: \textit{first-order approach} \cite{fcsc/ZhangLLG18,pr/BoutellLSB04, pr/ZhangZ07}, \textit{second-order approach} \cite{nips/ElisseeffW01, ml/furnkranz2008multilabel} and \textit{high-order approach} \cite{ml/ReadPHF11,tkde/TsoumakasKV11}. To solve MLL problems, the first-order approach decomposes MLL problems into a set of binary classification problems \cite{fcsc/ZhangLLG18,pr/BoutellLSB04}. However, these approaches ignore label correlations among labels, which play a crucial role in MLL \cite{tkde/ZhangZ14}. After realizing the importance of label correlation, more and more studies attempt to exploit it to improve MLL performance. Among them, the second-order approach considers the pairwise label correlations that refer to the relationship between two labels. The kind of these approaches generally transform MLL problems into bipartite ranking problems by enforcing that relevant labels should be ranked higher than irrelevant labels \cite{tkde/zhang2006multilabel,cvpr/li2017improving,ml/furnkranz2008multilabel}. Beyond second-order relationship, there exists more complex relationship between labels in many real-world scenarios. Therefore, many approaches begin to exploit high-order label correlations to handle the MLL problems recently \cite{tkdd/JiTYY10, ml/ReadPHF11,nips/GerychHBAR21, aaai/0002KBFG21}. For example, Zhao et al. \cite{aaai/0002KBFG21} leverage variational autoencoder to facilitate the learning process via exploiting high-order correlations among labels, while Wang et al. and Xun et al. \cite{icdm/WangDHHCF19, kdd/XunJSZ20} both design special neural network blocks to automatically extract label correlations to improve the label prediction performance. Although high-order approaches have the ability of stronger label correlation-modeling, they may suffer from high computational cost comparing to first and second-orders approaches \cite{2021Global}.

\subsection{Partial Multi-Label Learning}
Due to that the fully supervised data is difficult to collect, many reseachers tend to explore the weakly supervision data form to alleviate the heavy load of labeled data collection \cite{zhou2018brief}. PML is a recently emerging weakly supervised approch firstly proposed by Xie et al. \cite{aaai/XieH18}. In PML, each training instance is associated with a set of candidate labels that consist of \emph{relevant} labels and \emph{irrelevant} (noisy) labels and the goal is to learn a classifier assigning a set of labels accurately for unseen instances.

At the first glance, it seems that ML-CLL is an extreme case of PML, such that all PML methods are also applicable to ML-CLL. However, existing PML methods assume that noisy only composes a small portion in the candidate labels \cite{aaai/XieH20,aaai/SunFWLJ19,icdm/YuCDWLZW18,2021Global}, such that many approaches \cite{aaai/SunFWLJ19,icdm/YuCDWLZW18,2021Global} adopt matrix factorization matrix factorization to tackle PML problems, which decompose the candidate label matrix into the low-rank multi-label matrix and the sparse noisy label matrix. Compared to PML, the studied ML-CLL problem in this paper are target at the problem with only one complementary label, resulting in a high-noise PML problem on which the existing approaches can not be applicable. We will demonstrate the performance difference in the experimental part.

\subsection{Multi-Class Complementary Label Learning}
Currently, CLL problem is only considered in multi-class learning, whose goal is to predict a single relevant label per instance precisely from complementary labeled data. Previous approaches can be roughly grouped into two categories: (1) modeling the generative relationship between the complementary label and the relevant label \cite{nips/IshidaNHS17, icml/FengK000S20, icml/IshidaNMS19, eccv/yu_learning_2018,aaai/XuGCLZB20}; (2) modeling the probability of complementary labels from the learned discriminative classifier directly \cite{icml/GaoZ21,icml/Chou0LS20,ijcai/WangFZ21}.

The first multi-class CLL method belongs to category one. It models the generative relationship between complementary labels and relevant labels, and uses a such generative process to rewrite one-versus-all and pairwise comparison loss functions to derive an unbiased risk estimator \cite{nips/IshidaNHS17}. Ishida et al. \cite{icml/IshidaNMS19} realize that the method of \cite{nips/IshidaNHS17} is restricted to loss functions and propose a new method which can use arbitrary losses and models. A typical way to make use of the modeled generative process is through a transition matrix, which summarizes the probabilities of a label being complementary labels when relevant labels are given. Then, approaches apply a transition matrix to recover relevant labels from complementary labels \cite{eccv/yu_learning_2018,icml/IshidaNMS19,aaai/XuGCLZB20}. Compared with \cite{nips/IshidaNHS17, icml/IshidaNMS19}, transition matrix-based methods can map more complex generative relationship rather than uniform one only. Therefore, we tend to design a transition matrix-based method to solve ML-CLL problem with a different estimating way.

Differ from category one, approaches residing in category two directly model the probabilities of complementary labels from the learned classifier without the generative relationship \cite{icml/Chou0LS20, icml/GaoZ21,ijcai/WangFZ21}. Chou et al. propose a surrogate complementary loss framework based on complementary labels providing negative feedback during the training process \cite{icml/Chou0LS20}. Although its losses fail to derive an unbiased risk estimator, it achieves good performance on the multi-class CLL. In light of the property of the complementary label that the predictive probability of the complementary label is expected to approach zero, \cite{icml/GaoZ21} and \cite{ijcai/WangFZ21} propose a discriminative solution by directly modeling the probabilities of complementary labels from learned classifier to avoid the generative assumption. Due to that multi-class CLL approaches are designed for a single relevant label case, which are not suitable for the ML-CLL case that an instance is associated with multiple labels simultaneously. We will demonstrate that in the experimental part.

\section{Problem Setup}
\label{sec:setup}
In MLL, let $\mathcal X$ be the feature space and $\mathcal Y=\{l_1,l_2,\dots,l_K\}$ be the finite label space with $K$ possible class labels ($K>2$). A multi-label instance $\bm x\in\mathcal X$ is equipped with a set of relevant labels $Y\subseteq\mathcal Y$. $(\bm x, Y)$ is independently sampled from an unknown joint probability distribution $p(\bm x, Y)$. Here we exclude the special cases of $Y = \emptyset$ nor $\mathcal Y$ to ensure relevant labels and complementary labels both exist. For convenience, we use a binary vector $\bm y=[y^1,y^2,\dots,y^K]\in\{0,1\}^K$ to denote $Y$, where $y^k=1$ indicates that $l_k\in Y$ is relevant to $\bm x$ and $0$ otherwise. Suppose $D=\{(\bm x_i, \bm y_i)\}^n_{i=1} \stackrel{\text { i.i.d. }}{\sim} p(\bm x, Y)$ is the training set with $n$ instances. The goal of MLL is to learn a multi-label classifier $h: \mathcal X\rightarrow 2^{\mathcal Y}$, which can predict a set of relevant labels for any unseen instance. Instead of learning $h$ directly, most MLL methods tend to learn a real-valued decision function $\bm f:\mathcal X\rightarrow\mathbb{R}^K$ via minimizing the expected risk
\begin{align}
	R_{L}(\bm f)=\mathbb E_{p(\bm x,Y)}[L(\bm f(\bm x),\bm y)], \label{eq_1}
\end{align}  
where $L$ is a proper MLL loss function \cite{aaai/0002KBFG21}, such as BCE loss. $\bm f(\bm x)$ is usually interpreted as a probability vector: $f^k(\bm x)$ is the $k$-th entry of $\bm f(\bm x)$ and predicts the confidence score that label $l_k$ is relevant to $\bm x$, i.e., if properly normalized then $p(y^k = 1|\bm x)$. Due to that $p(\bm x, Y)$ is unknown, the expected risk is usually approximated by the empirical risk $\widehat R_L(\bm f)=\frac{1}{n}\sum_{i=1}^n L(\bm f(\bm x_i),\bm y_i)$. If denoting the optimal classifier learned from the expected risk as $\bm f^*$, i.e., $\bm f^*=\mathrm{argmin}_{\bm f}\;R_{L}(\bm f)$, then $\widehat{\bm f}^*$ denotes the optimal classifier learned by minimizing the empirical risk, i.e., $\widehat{\bm f}^*=\mathrm{argmin}_{\bm f}\;\widehat R_L(\bm f)$. 

In ML-CLL studied in this paper, each training instance is equipped with a \emph{single} complementary label. The complementary labeled instance $(\bm x, \bar y)\in (\mathcal{X}, \mathcal{Y})$ is drawn from an unknown joint probability distribution $p(\bm x, \bar y)$, where $\bar y\in\mathcal Y\setminus Y$ is a complementary label of $\bm x$.  $\bar y$ can be presented as a $K$-dimensional vector $\bm{\bar y}=[\bar y^1,\bar y^2,\dots,\bar y^K]$. If label $l_j$ is selected as the complementary label to $\bm x$ ($\bar y=l_j$), then $\bar y^j$ is one and all other elements are zero in $\bm{\bar y}$. We utilize $\widehat Y=\mathcal Y\setminus\bar y$ to denote \emph{the candidate label set} of $\bm x$. Let a $K$-dimension vector  $\bm{\widehat y}=[\widehat y^1,\widehat y^2,\dots,\widehat y^K]$ to be the corresponding vector representation of subset $\widehat Y$, where all elements are one except that the one  corresponding to the complementary label is set to be zero ($\bm{\widehat y} = \bm{1} -\bm{\bar y}$).

Let $\bar D=\{(\bm x_i,\bar y_i)\}_{i=1}^n \stackrel{\text { i.i.d. }}{\sim} p(\bm x, \bar y)$ be the ML-CLL training set with $n$ instances. The expected risk of multi-labeled CLL is defined over $p(\bm x, \bar y)$:
\begin{align}
	R_{\bar L}(\bm f)=\mathbb E_{p(\bm x, \bar y)}[\bar L(\bm f(\bm x),\bm{\bar y})], \label{eq_2}
\end{align} 
where $\bar L$ denotes a ML-CLL loss, which will be proposed later this paper. Similarly, the corresponding empirical risk is described as $\widehat R_{\bar L}(\bm f)=\frac{1}{n}\sum_{i=1}^n \bar L(\bm f(\bm x_i),\bm{\bar y}_i)$.

\section{The Proposed Approach}
\label{sec:approach}
In this section, we first introduce the definition of the transition matrix in MLL and analyze why the estimated transition matrix using multi-class techniques is unsuitable for ML-CLL. Then, we describe an advanced two-step way to estimate the transition matrix in the MLL case. Finally, we prove our approach is classifier-consistent with a mild assumption.

%

\subsection{Transition Matrix for ML-CLL}
In ML-CLL, we start by introducing a transition matrix $\mathbf{\tilde{T}}$ that summarizes the probabilities for a complementary label given a set of relevant labels. More specifically, the transition matrix $\mathbf{\tilde{T}}$ is defined as $\mathbf{\tilde{T}}_{kj}=p(\bar y^j=1|Y=C_k)$ where $C_k\in \mathcal{Y}'=\{2^{\mathcal Y}-\emptyset-\mathcal Y\}$ ($k\in [2^K-2]$) is the $k$-th label subset. If $l_j\in C_k$, then $\mathbf{\tilde{T}}_{kj}=0$ because the label $l_j$ has no chance to be selected as the complementary label. In this paper, we employ the same class-dependent assumption as the multi-class CLL approach \cite{eccv/yu_learning_2018}: $p(\bar y|Y,\bm x)=p(\bar y|Y)$ as $\bar y$ and $\bm x$ are conditionally independent given $Y$. Then we can obtain the following equation: 
\begin{align}
	\label{eq_3}
	p(\bar y^j=1|\bm x)=\sum_{C\in \mathcal{Y}', l_j\notin C} p(\bar y^j=1|Y=C)p(Y=C|\bm x),
\end{align} 
where we assume the label $l_j$ is a complementary label of $\bm x$. Then, according to Eq.(\ref{eq_3}), $p(\bar y|\bm x)$ can be approximated by $p(Y|\bm x)$ when the transition matrix $\mathbf{\tilde{T}}$ is known. If considering all possible label subsets of $\mathcal Y'$ as $C$, we have $\mathbf{\tilde{T}}\in \mathbb{R}^{(2^K-2)\times K}$, i.e., the size of $\mathbf{\tilde{T}}$ depends on the size of the power set of $\mathcal{Y}'$. Practically, the power set of $\mathcal Y'$ would be computationally prohibitive and even impossible to store, since $2^K-2$ is an extremely large number when the number of possible labels $K$ is large. To solve this combinatorial explosion problem, we explore a more practical way to use an alternative lower-dimensional transition matrix to replace the higher-dimensional one. We start investigating the feasibility of the alternative lower-dimensional matrix from Theorem \ref{theo:matrix}.
\begin{theorem}
	\label{theo:matrix}
	Given an instance $\bm x$, suppose $Y$ is the relevant label set and the label $l_j$ is the complementary label which is randomly selected. Then the following equality holds: 
	\begin{align*}
		p(\bar y^j=1|\bm x)=\sum_{C\in \mathcal{Y}', l_j\notin C} p(\bar y^j=1| Y=C)p(Y=C|\bm x)\\
		\ge \sum_{k=1, k\neq j}^{K} p(\bar y^j=1|y^k=1)p(y^k=1|\bm x).
	\end{align*}
\end{theorem}

\noindent The second inequality holds because of addition rule of probability. The detailed proof is in Appendix \ref{apx:theorem_1}. Theorem \ref{theo:matrix} shows that using $\mathbf T$ to approximate $p(\bm{\bar y}|\bm x)$ is a lower bound of using $\mathbf{\tilde{T}}$ to approximate $p(\bm{\bar y}|\bm x)$. Observed by Eq.(\ref{eq_3}), we find that our main goal transforms from precisely predicting the relevant label set $Y$ of $\bm x$ to precisely predicting its complementary label $\bar y$ via the transition matrix $\mathbf{\tilde{T}}$. This means that we need to maximize the predictive probability of the complementary label of $\bm x$, i.e., maximizing $p(\bar y|\bm x)$. From this point of view, Theorem \ref{theo:matrix} theoretically shows the feasibility of using a low-dimension transition matrix to replace the high-dimension $\mathbf{\tilde{T}}$, because we optimize by maximizing the lower bound of Eq.(\ref{eq_3}). Let $\mathbf T\in [0,1]^{K\times K}$  denote the lower-dimensional transition matrix, where the $(k, j$)-th element of $\mathbf T$ is $\mathbf T_{kj}=p(\bar y^j=1|y^k=1)$, and $\mathbf T_{kj}=0$ when $k=j$. Thus, we adopt the $K\times K$ matrix $\mathbf T$ as the transition matrix in the following of the paper to avoid the pain in computation and storage brought up by the $(2^K-2)\times K$ matrix $\mathbf{\tilde{T}}$.

\subsection{Distortion in Estimating the Transition Matrix}
\label{sec:distortion}
Before exploring how the transition matrix estimated by multi-class CLL is distorted from that of ML-CLL, we first introduce the transition matrix estimated by multi-class CLL techniques. Suppose $\mathbf Q\in [0,1]^{K\times K}$ be the transition matrix estimated in multi-class CLL. Recalling the approach \cite{eccv/yu_learning_2018}, it  estimates the transition matrix under a special assumption: for each label $l_k$, existing an anchor set $\mathcal{S}_{\bm x|l_k}\subset \mathcal{X}$ such that $p(y^k=1|\bm x)=1$ and $p(y^{k'}=1|\bm x)=0$ ($l_{k'}\in\mathcal{Y}\setminus\{l_k\}$). With this assumption and regardless of label correlations, the estimation of $\mathbf Q_{kj}$ is $p(\bar y^j=1|y^k=1)=p(\bar y^j=1|\bm x)$ iff $\bm x$ is sampled from $\mathcal{S}_{\bm x|l_k}$, where $\mathbf Q_{kj}$ is the $k$-th row and $j$-th column element of $\mathbf Q$.

To measure the distortion between $\mathbf T$ calculated in ML-CLL and the estimated $\mathbf Q$, we define their difference on the complementary label $l_j$ of $\bm x$ as follows
\begin{align}
	\label{eq_4}
	\ell_j=\sum_{k=1}^K |\mathbf T_{kj}-\mathbf Q_{kj}|. 
\end{align}
The larger value of $\sum_{j=1}^K \ell_j$ indicates that $\mathbf T$ deviates further from $\mathbf Q$. As we know, label correlations and co-occurred multiple labels are key properties of MLL. Due to that the correlations among labels are intricate, directly calculating $\mathbf T$ from all label correlations will bring high computational cost. For convenience, we give a simple case of MLL including label correlations -- at most two labels can co-occur for an instance, and the rest of labels are mutually exclusive -- to facilitate us calculating $\mathbf T$ from label correlations and explore the distortion of $\mathbf T$ and $\mathbf Q$. We start to study the above contents from the definition of mutually exclusive.

\begin{definition}
	For any $\bm x \in\mathcal X$, only a label is relevant to $\bm x$, i.e. $|Y|=1$, which labels are mutually exclusive.
\end{definition}

\noindent Under the simple case in MLL, in Theorem \ref{theo_T}, we state how to estimate $\mathbf T$ directly from label correlations, and the distortion of $\mathbf T$ and $\mathbf Q$. 

\begin{theorem}
	\label{theo_T}
	Under a MLL scenario: suppose the labels $l_{z_1}, l_{z_2}\in\mathcal Y$ ($z_1, z_2 \in [K], z_1\neq z_2$) are dependent, and the labels belonging to $\mathcal Y\setminus\{l_{z_1},l_{z_2}\}$ are mutually exclusive. For any $\bm x$, its label set $Y \subseteq \{l_{z_1},l_{z_2}\}$ and $Y \neq \emptyset$. Let the label $l_j$ ($j\in[K], j\neq z_1,z_2$) be the complementary label of $\bm x \in \mathcal{X}$. $\mathbf T_{z_1j}$ and $\mathbf T_{z_2j}$ calculated from label correlations satisfy
	\begin{align*}
		\mathbf T_{z_1j}=\frac{p(\bar y^j=1|\bm x)}{p(y^{z_2}=1|\bar y^j=1,y^{z_1}=1, \bm x)p(y^{z_1}=1|\bm x)}, 
		\\
		\mathbf T_{z_2j}=\frac{p(\bar y^j=1|\bm x)}{p(y^{z_1}=1|\bar y^j=1,y^{z_2}=1, \bm x)p(y^{z_2}=1|\bm x)},
	\end{align*}
	where $[K]$ denotes the integer set $\{1,2,\dots,K\}$. The difference of $\mathbf{T}$ and $\mathbf{Q}$ on the complementary label $l_j$ is
	\begin{align*}
		\ell_j \geq 2(\frac{1}{\xi^2}-1)p(\bar y^j=1|\bm x),
	\end{align*}
	where $\xi=\max \{ p(y^{z_2}=1|\bar y^j=1,y^{z_1}=1, \bm x),p(y^{z_1}=1|\bar y^j=1,y^{z_2}=1, \bm x) \}$.
\end{theorem}
\noindent The proof is provided in Appendix \ref{apx:estimate}. From Theorem \ref{theo_T}, we can see that calculating the transition matrix from label correlations is more complex than estimating one without label correlations, and the relevant label sets of instances need to be known. Moreover, Theorem \ref{theo_T} shows that there is a distortion between $\mathbf{T}$ and $\mathbf{Q}$, which widely exists in multi-labeled cases since each multi-label instance is relevant to multiple labels. The above learning scenario only considers the pairwise label correlations, while there exists a more complex relationship among labels. Similarly, under a realizable computational cost, we construct another simple MLL scenario with more complex label relationships to explore factors that affect $\ell_j$ in Corollary \ref{coro}.

\begin{corollary}
	\label{coro}
	Under a MLL scenario: there are $m$ ($m\geq 2$) labels $l_{z_1}, l_{z_2},\dots,l_{z_m} \in \mathcal Y$ $(z_1,\dots, z_m\in [K])$ that are dependent, while the labels belong to $\mathcal Y\setminus\{l_{z_1},l_{z_2}, \dots, l_{z_m}\}$ are mutually exclusive. For any $\bm x \in\mathcal{X}$, its relevant set $Y \subseteq \{l_{z_1}, l_{z_2}\dots, l_{z_m}\}$ and $Y \neq \emptyset$. Suppose the label $l_j$ is the complementary label of $\bm x$. The difference $\ell_j$ between $\mathbf{T}$ and $\mathbf{Q}$ has
	\begin{align*}
		\ell_j\geq m(\frac{1}{\xi^{m}}-1)p(\bar y^j=1|\bm x),
	\end{align*}
	where $\xi = \mathrm{max}\{p(y^{z_m}=1|\bar y^j=1,y^{z_1}=1,\dots,y^{z_{m-1}}=1, \bm x), p(y^{z_{m-1}}=1|\bar y^j=1,y^{z_1}=1,\dots,y^{z_{m-2}}=1, y^{z_{m}}=1, \bm x),\dots, p(y^{z_1}=1|\bar y^j=1,y^{z_2}=1,\dots,y^{z_{m}}=1, \bm x)\}$ $(\xi\in(0,1])$.
\end{corollary}
\noindent The proof is shown in Appendix \ref{apx:cor}. According to Corollary \ref{coro}, when label correlations are more complex, the distortion of the transition matrix estimated by the multi-class CLL approach is more serious as $m$ increases. Meanwhile, it demonstrates that the ML-CLL problem cannot be solved by current techniques in multi-class CLL. 

\subsection{Estimation $\mathbf T$ with Label Correlations}
As discussed above, calculating the transition matrix $\mathbf T$ from label correlations needs instances whose relevant label sets are known. Moreover, calculating $\mathbf T$ is more and more difficult as relationships among labels become more complex by observing the results of $\mathbf T$ in Theorem \ref{theo_T} and Corollary \ref{coro}. Due to that multi-labeled data are unavailable for our setting, we propose a two-step method to estimate $\mathbf T$ from candidate labels, and it can reduce the complexities in calculating $\mathbf T$ from label correlations. This two-step method includes: (1) computing an initial transition matrix $\mathbf S\in[0,1]^{K\times K}$ from candidate labels by decomposing the multi-label problem into a series of binary classification problem; (2) obtaining the final estimation of $\mathbf T$ by using label correlations to correct $S$. 

\textbf{Computing an initial transition matrix $\mathbf S$.} Let $\mathbf S_{kj}=p(\bar y^j=1|\widehat y^k=1)$ be an initial transition probability, which is a $(k, j)$-th element of $\mathbf S$. We caulculate $\mathbf S$ from candidate labels of instances. Multiplication theorem of probability \footnote{$p(\bm x,\bar y^j=1,\widehat y^k=1)=p(\bar y^j=1|\widehat y^k=1, \bm x)p(\bm x|\widehat y^k=1)p(\widehat y^k=1)=p(\bar y^j=1|\widehat y^k=1)p(\bm x|\bar y^j=1,\widehat y^k=1)p(\widehat y^k=1) \Rightarrow p(\bar y^j=1|\widehat y^k=1, \bm x)p(\bm x|\widehat y^k=1)=p(\bar y^j=1|\widehat y^k=1)p(\bm x|\bar y^j=1,\widehat y^k=1)$} is applied to calculate $\mathbf S_{kj}$ and ensure that the following equation holds:
\begin{align}
	\label{eq_5}
	\mathbf S_{kj}&=p(\bar y^j=1|\widehat y^k=1)
	\\ 
	&=p(\bar y^j=1|\widehat y^k=1) \int p(\bm x|\bar y^j=1,\widehat y^k=1)d\bm x \notag
	\\
	&=\int p(\bar y^j=1|\widehat y^k=1,\bm x)p(\bm x|\widehat y^k=1)d\bm x  \notag
	\\
	&=\mathbb{E}_{p(\bm x|\widehat y^k=1)}[p(\bar y^j=1|\widehat y^k=1,\bm x)], \notag
\end{align}
where $j,k\in[K]$ and $j\neq k$. In practice, $\mathbb{E}_{p(\bm x|\widehat y^k=1)}[p(\bar y^j=1|\widehat y^k=1,\bm x)]$ can be approximated by the expectation of $p(\bar y^j=1|\widehat y^k=1,\bm x)$ over the conditional distribution $p(\bm x|\widehat y^k=1)$. Assuming $\bar y$ and $\widehat Y$ are conditionally independent given $\bm x$, so $p(\bar y^j=1|\widehat y^k=1,\bm x)=p(\bar y^j=1|\bm x)$. Intuitively, $p(\bar y^j=1|\bm x)$ can be approximated by the classifier learned from $\bar D$ to predict the probability of complementary labels. Let $A_k$ denote the subset of $\bm x$ in $\bar D$ with $\widehat y^k=1$, which satisfies the conditional distribution $p(\bm x|\widehat y^k=1)$. Thus, $\mathbf S_{kj}$ can be estimated by
\begin{align}
	\label{eq_6}
	\mathbf S_{kj}&=\frac{1}{|A_k|}\sum_{\bm x\in A_k} p(\bar y^j=1|\widehat y^k=1, \bm x)
	\\ &=\frac{1}{|A_k|}\sum_{\bm x\in A_k} p(\bar y^j=1|\bm x).  \notag
\end{align}

\begin{figure}
	\centering
	\centering
	\includegraphics[width=0.9\columnwidth]{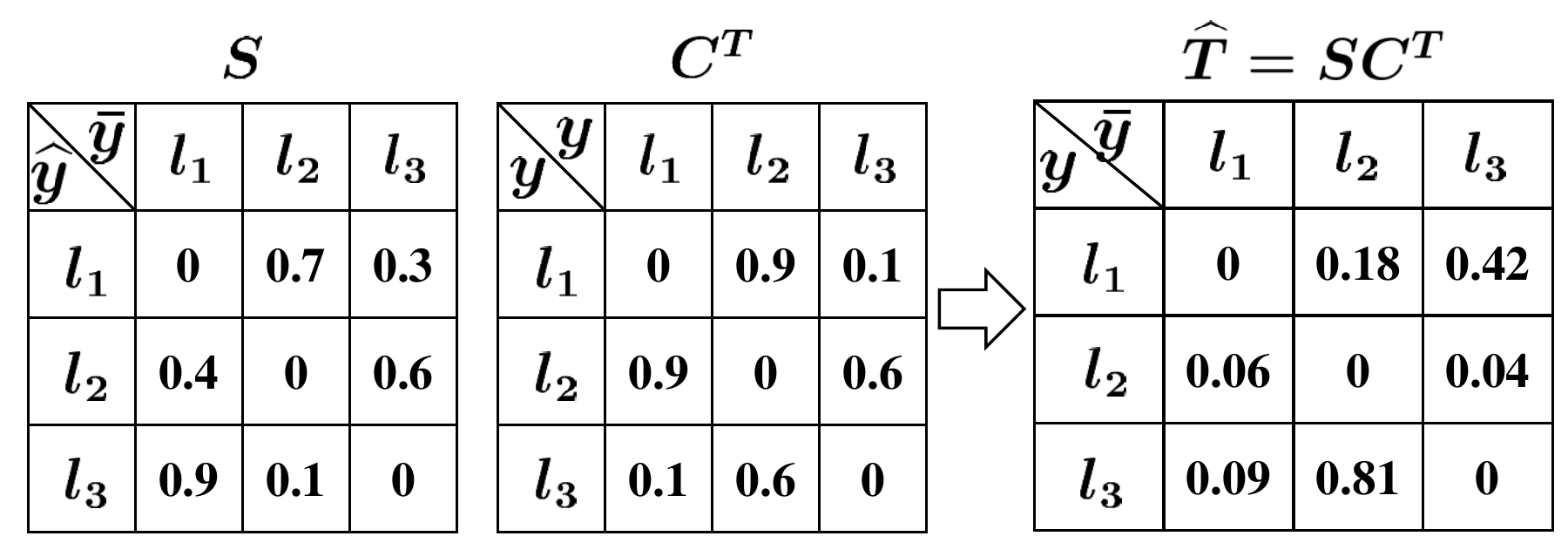}
	\vskip -0.1in
	\caption{An example of correcting $S$ with label correlations.}
	\label{fig_2}
	\vskip -0.1in
\end{figure}

\textbf{Estimating $\mathbf T$ with label correlations.} The calculating procedure of $\mathbf S$ lacks exactly supervised data. Observed by the transition probabilities of $\mathbf{T}$ calculated from label correlations in subsection \ref{sec:distortion}, we can find that they are affected by label correlations. Moreover, a label that is low-co-occurred to the relevant labels could be preferentially selected as the complementary label from the view of label correlations. For example, considering \textit{water} as the relevant label; in this case, \textit{desert} (low-co-occurred label) will have a larger chance to be selected as the complementary label compared to \textit{fish} (high-co-occurred label). Motivated by these findings, we use label correlations to correct the initial matrix $\mathbf S$ to estimate $\mathbf T$ by enforcing the addition of relationships among labels.

Suppose $\mathbf C\in[0,1]^{K\times K}$ be a label correlation matrix, where the element $\mathbf C_{kj}$ represents the correlation between labels $l_k$ and $l_j$. The value of $\mathbf C_{kj}$ is larger when the correlation of labels $l_k$ and $l_j$ is stronger. Following \cite{aaai/XieH18, pci/DiplarisTMV05}, we adopt the co-occurrence rate of two candidate labels as their correlations. Finally, the transition matrix $\mathbf T$ can be estimated by $\mathbf {\widehat T} =\mathbf S\mathbf C^T$, where $\mathbf{\widehat T}_{kj}=0$ if $k=j$, and normalizing $\mathbf T$ by row. 

Fig. \ref{fig_2} is an example of refining procedure. As can be seen from the Fig. \ref{fig_2}, though the estimated initial probability of $p(\bar y^2=1|\widehat y^1=1)$ is higher than $p(\bar y^3=1|\widehat y^1=1)$ in $\mathbf S$, the value of $p(\bar y^2=1|y^1=1)$ is lower than $p(\bar y^3=1|y^1=1)$ in $\mathbf{\widehat T}$. This is because the labels $l_1$ and $l_2$ have a strong correlation as shown in $\mathbf C$, so the label $l_2$ has a lower chance to be selected as the complementary label for the label $l_1$. The corrected initial transition matrix $\mathbf{S}$ agrees with our expectation on the low-co-occurred labels that tend to be selected as complementary labels preferentially. In practice, the estimation of $\mathbf T$ depends on $p(\bar y|\bm x)$, where the classifier should perfectly model the probability of complementary labels. When data equipped with complementary labels is sufficiently, the perfect model is capable of modeling $p(\bar y|\bm x)$.

\vspace{-0.05in}
\subsection{A Classifier-Consistent Approach}
According to the transition matrix $\mathbf T$, we can derive the probability of complementary labels from multi-label classifier. Let $\bm{\bar f}(\bm x) \in\mathbb R^K$ be a complementary label classifier, which is defined as 
\begin{align}
	\label{eq_8}
	\bm{\bar f}(\bm x)=\mathbf T^T\bm f(\bm x), 
\end{align}
where $\bm{\bar f}(\bm x)$ is applied to approximate $p(\bm{\bar y}|\bm x)$, $\bar f^j(\bm x)$ refers to the $j$-th element of $\bm{\bar f}(\bm x)$. ML-CLL problems aim to recover a set of relevant labels per instance from a complementary label. Since training instances are associated with complementary labels, the common loss functions of MLL are unsuitable for ML-CLL. Therefore, we define a complementary loss function $\bar L$ as 
\begin{align}
	\label{eq_9}
	\bar L(\bm f(\bm x),\bm{\bar y})=L(\bm{\bar f}(\bm x),\bm{\bar y})=L(\mathbf T^T\bm f(\bm x),\bm{\bar y}).
\end{align}

Denote by $\bm f_{CL}^*$ the minimizer of $R_{\bar L}(\bm f)$, the minimizer $\widehat{\bm f}^*_{CL}$ of $\widehat R_{\bar L}(\bm f)$ is used to approximated $\bm f_{CL}^*$. Recalling the definition of classifier-consistent, if a classifier learned by an approach finally converges to the optimal classifier $\bm f^*$ learned in MLL as the number of instances increases, then this approach is classifier-consistent \cite{cvpr/PatriniRMNQ17,nips/XiaLW00NS19,icml/LvXF0GS20}. We derive our proposal is classifier-consistent based on a mild assumption:
\begin{assumption}
	\label{assumption}
	Suppose the transition matrix $\mathbf T$ is invertible and can perfectly recover the relationship between relevant labels of $\bm x$ and its complementary label. Then, we have $\bm{\bar y}=\mathbf T^T \bm y$.
\end{assumption}

\begin{algorithm}[t]
	\caption{MLCL Algorithm}
	\LinesNumbered
	\KwIn{\\
		$\bar D$: the complementary-label training set $\{(\bm x_i, \bm{\bar{y}_i)} \}_{i=1}^n $\;
		$E$: the number of epochs\; 
		$ \mathcal{A} $: an external stochastic optimization algorithm;
	}
	\KwOut{\\
		$\theta$: model parameter for {$\bm f(\bm x;\theta)$};
	}
	\If{$\mathbf T$ is unknown}{
		Train a classifier $\bm{\bar f}(\bm x)$ with the $softmax$ output layer and Cross-Entropy loss on $\bar D$\;
		Fill $\mathbf S\in[0,1]^{K\times K}$ with zeros\;
		\For{$k=1$ to $K$}{
			num = 0\;
			\For{$(\bm x_i,\bm{\bar y_i})\in\bar D$ such that $\bar y_i^k=0$}{
				num += 1\;
				$\mathbf S_{k\cdot}+=\bm{\bar f}(\bm x_i)$;  $//$add $\bm{\bar f}(\bm x_i)$ to $k$-th row of $\mathbf S$
			} 
			$\mathbf S_{k\cdot}/=$ num;
		} 
		$\widehat{ \mathbf T}=\mathbf S\mathbf C^T$;
	} 
	\For{$ t=1 $ to $ E $}{
		Let $\mathcal L$ be the risk, $\mathcal L=\frac{1}{n}\sum_{i=1}^n\mathcal{\bar L}(\bm f(\bm x_i),\bm{\bar y_i})=\frac{1}{n}\sum_{i=1}^n ( L( \widehat{ \mathbf T}^T\bm f(\bm x_i),\bm{\bar y_i})+\left\| \bm{\bar y_i}- \widehat{ \mathbf T}^T\bm f(\bm x_i)\right\|_F^2)$\;
		Set gradient $ -\nabla_\theta \mathcal{L} $\;
		Update $ \theta $ by $ \mathcal{A} $;
	}
\end{algorithm}

\noindent With Assumption \ref{assumption}, our approach trained on $\bar L$ can be inferred to be classifier-consistent, which is stated in Theorem \ref{consist}. Naturally, Theorem \ref{consist} guarantees that the optimal classifier learned from complementary labeled data converges to the optimal one learned from fully supervised MLL.
\begin{theorem}
	\label{consist}
	With Assumption \ref{assumption}, suppose the transition matrix $\mathbf T$ is invertible, then the ML-CLL optimal classifier $\bm f_{CL}^*$ converges to the MLL optimal classifier $\bm f^*$, i.e., $\bm f_{CL}^*=\bm f^*$.
\end{theorem}
\noindent The proof is represented in Appendix \ref{apx:consist}. Thanks to BCE loss is a popular loss function in MLL, we adopt BCE loss as the base in this paper, then $\bar L$ is expressed as
\begin{align}
	\label{eq_10}
	\bar L(\bm f(\bm x),\bm{\bar y})=&-\bm{\bar y}\mathrm{log}(\mathbf{T}^T \bm f(\bm x))-(\bm 1-\bm{\bar y})\mathrm{log}(\bm 1-\mathbf{T}^T \bm f(\bm x))),
\end{align}
where $\bm 1$ denotes a $K$-dimensional vector with 1 for all elements.

\section{Regularization-Based Enhancement}
\label{sec:mse}
In this section, an MSE-based regularization of our approach is described. And we attempt to combine a small amount of relevant labels to explore more possibilities of our proposal.

\vspace{-0.1in}
\subsection{An MSE-Based Regularization}
Previous works indicate that CE loss always makes the model focus on hard instances that are difficult to be classified precisely, while MSE loss and Mean Absolute Error (MAE) loss are less sensitive to hard instances since they treat per instance coequally \cite{nips/ZhangS18,aaai/GhoshKS17}. As this property, the convergence rate of CE loss is superior to MSE loss and MAE loss, whereas this property makes CE loss more prone to the overfitting problem than MSE loss and MAE loss when noisy labels present at training data \cite{nips/ZhangS18,aaai/GhoshKS17}. Actually, an excellent approach can converge quickly during the training process, and shows good generalization ability and robustness for unseen instances\cite{ijcai/WangFZ21}.

Obviously, BCE loss has a similar property to CE loss, which results in an excellent convergence rate of approaches. Meanwhile, approaches based on BCE loss are easy to suffer from the overfitting problem when using noisy labeled data to learn. In fact, ML-CLL is a problem setting with dense noisy labels, BCE loss may cause the overfitting problem of a model in ML-CLL. To cope with this problem, we introduce an MSE-based regularizer based on MSE loss (i.e. $\ell_2$-norm regularization) to balance the robust and convergence requirement of the proposed approach. Hence, the MSE-based regularizer is defined as:
\begin{align}
	\label{eq_11}
	\bar L_{mse}(\bm f(\bm x),\bm{\bar y})=\left\| \bm{\bar y}-\mathbf T^T\bm f(\bm x)\right\|^2_F.
\end{align}

Finally, we combine the complementary loss and the MSE-based regularizer term, which leads to our target loss:
\vspace{-0.05in}
\begin{align}
	\label{eq_12}
	\mathcal{\bar L}(\bm f(\bm x),\bm{\bar y})=\bar L(\bm f(\bm x),\bm{\bar y})+\beta\bar L_{mse}(\bm f(\bm x),\bm{\bar y}),
\end{align}
where $\beta$ is the trade-off parameter and set as 1 (the selection shown in Section \ref{sec:experiments}). The all procedure of the proposed approach (called MLCL) is shown in Algorithm 1.

\subsection{Incorporation of Relevant Labels}
\label{sec:rl}
In many practical situations, we can use complementary labels and relevant labels to learn more accurate classifiers, which is highly practical implementation. To this end, motivated by \cite{nips/IshidaNHS17, acml/KatsuraU20}, let us design a reasonable combination of the loss derived from complementary labeled data and relevant labeled data:
\begin{align}
	\label{eq_13}
	\mathcal{\tilde{L}}(\bm f(\bm x),\bm{\bar y}, \bm{\tilde y})=\bar L(\bm f(\bm x),\bm{\bar y})+\left\| \bm{\tilde y}-\bm f(\bm x)\right\|^2_F,
\end{align}
where $\bm{\tilde y}=[\tilde y^1, \dots, \tilde y^1]\in\{0, 1\}^K$ denotes a binary vector of relevant labels $\tilde Y$ of $\bm x$, in which $\tilde y^1=1$ when the label $l_k\in \tilde{Y}$. To provide more practicability, we do not restrict given relevant labels $\tilde Y$ to must be equal to the set of relevant labels $Y$, which means $\tilde Y\subseteq Y$ and $\tilde Y\neq \emptyset$. 

As explained in the instruction, we can naturally collect data associated with complementary labels and relevant labels via crowdsourcing \cite{sindlinger2010crowdsourcing}. Our loss function Eq.(\ref{eq_13}) can leverage both kinds of labeled data to learn better classifiers. We will experimentally show the usefulness of this combination method in Section \ref{sec:experiments}.

\begin{table}[t]
	\caption{Statistics of datasets.}
	\label{tab_data}
	\vspace{-0.15in}
	\begin{center}
		\begin{small}
			\begin{tabular}{lcccc}
				\hline
				Datasets   & $|\mathcal S|$ & $dim(\mathcal S)$ & $L(\mathcal S)$ & $LCard(\mathcal S)$\\ \hline
				scene      & 2407      & 294     & 6   &   1.07       \\
				yeast      & 2417      & 103     & 14    & 4.23       \\
				eurlex\_dc & 8636      & 5000    & 15  & 1.02         \\
				eurlex\_sm & 13270     & 5000    & 15  & 1.74         \\
				corel5k    & 4194      & 499     & 15   & 1.70        \\
				corel16k   & 11103     & 120     & 15       & 1.77    \\
				bookmark   & 38912     & 2150    & 15      & 1.25     \\
				delicious  & 14784     & 500     & 15    & 4.32       \\ \hline
			\end{tabular}
		\end{small}
	\end{center}
	\vspace{-0.1in}
\end{table} 

\begin{table*}[t]
	\caption{Experimental results (mean ± std) on training data with uniform complementary labels. The best performance of each dataset is presented in \textbf{boldface}, where $\bullet/\circ$ indicates whether MLCL is superior/inferior to baselines (with 5\% t-test).}
	\renewcommand\arraystretch{1.2}
	\label{tab_2}
	\vspace{-0.25in}
	\begin{center}
		\setlength{\tabcolsep}{5mm}{
			\begin{tabular}{llllllll}
				\hline
				\multicolumn{1}{c}{Methods}                       & \multicolumn{1}{c}{ML-KNN} & \multicolumn{1}{c}{LIFT} & \multicolumn{1}{c}{fpml} & \multicolumn{1}{c}{PML-lc}                 & \multicolumn{1}{c}{PML-LRS}                & \multicolumn{1}{c|}{L-UW}               & MLCL                                \\ 
				\hline
				\multicolumn{8}{c}{Ranking loss $\downarrow$}    \\ 
				\hline
				scene                                              & .340±.032$\bullet$         & .289±.020$\bullet$       & .504±.025$\bullet$       & .490±.025$\bullet$                         & \textbf{.258±.007$\circ$} & \multicolumn{1}{l|}{.372±.028$\bullet$} & .259±.030                           \\
				yeast                                              & .247±.012$\bullet$         & .298±.012$\bullet$       & .233±.013$\bullet$       & .251±.015$\bullet$                         & .464±.019$\bullet$                         & \multicolumn{1}{l|}{.214±.011}          & \textbf{.211±.013} \\
				eurlex\_dc                                         & .303±.016$\bullet$         & .286±.016$\bullet$       & .488±.033$\bullet$       & .347±.025$\bullet$                         & .316±.011$\bullet$                         & \multicolumn{1}{l|}{.598±.024$\bullet$} & \textbf{.229±.026} \\
				eurlex\_sm                                         & .336±.010$\bullet$         & .346±.012$\bullet$       & .488±.006$\bullet$       & .436±.011$\bullet$                         & .332±.009$\bullet$                         & \multicolumn{1}{l|}{.646±.015$\bullet$} & \textbf{.312±.014} \\
				corel5k                                            & .379±.034                  & .433±.037$\bullet$       & .444±.026$\bullet$       & .406±.075$\bullet$                         & \textbf{.334±.009$\circ$} & \multicolumn{1}{l|}{.367±.031}          & .349±.035                           \\
				corel16k                                           & .328±.047                  & .392±.027$\bullet$       & .420±.033$\bullet$       & .457±.046$\bullet$                         & .303±.005                                  & \multicolumn{1}{l|}{.303±.035}          & \textbf{.289±.042} \\
				bookmark                                           & .384±.006$\bullet$         & .310±.007$\bullet$       & .469±.019$\bullet$       & .454±.036$\bullet$                         & .260±.004                                  & \multicolumn{1}{l|}{.303±.010$\bullet$} & \textbf{.252±.013} \\
				delicious                                          & .398±.004$\bullet$         & .383±.003$\bullet$       & .438±.008$\bullet$       & .445±.015$\bullet$                         & .305±.002                                  & \multicolumn{1}{l|}{.302±.006}          & .310±.003                           \\
				\hline
				\multicolumn{8}{c}{One Error $\downarrow$}    \\ \hline
				scene                                              & .692±.030$\bullet$         & .605±.023$\bullet$       & .815±.027$\bullet$       & .717±.021$\bullet$                         & .540±.023$\bullet$                         & \multicolumn{1}{l|}{.609±.041$\bullet$} & \textbf{.427±.018} \\
				yeast                                              & .297±.029$\bullet$         & .284±.028$\bullet$       & .251±.025                & .583±.026$\bullet$                         & .738±.102$\bullet$                         & \multicolumn{1}{l|}{.251±.025}          & \textbf{.251±.023} \\
				eurlex\_dc                                         & .776±.031$\bullet$         & .670±.013$\bullet$       & .925±.016$\bullet$       & .774±.015$\bullet$                         & .847±.010$\bullet$                         & \multicolumn{1}{l|}{.837±.034$\bullet$} & \textbf{.594±.035} \\
				eurlex\_sm                                         & .689±.012$\bullet$         & .679±.009$\bullet$       & .872±.011$\bullet$       & .662±.012                                  & .731±.005$\bullet$                         & \multicolumn{1}{l|}{.696±.008$\bullet$} & \textbf{.656±.029} \\
				corel5k                                            & .815±.048$\bullet$         & .842±.056$\bullet$       & .854±.035$\bullet$       & .811±.062$\bullet$                         & .756±.010                                  & \multicolumn{1}{l|}{.769±.034}          & \textbf{.736±.065} \\
				corel16k                                           & .736±.056                  & .789±.046$\bullet$       & .816±.025$\bullet$       & .946±.028$\bullet$                         & .730±.000$\bullet$                         & \multicolumn{1}{l|}{.693±.057}          & \textbf{.690±.056} \\
				bookmark                                           & .801±.006$\bullet$         & .649±.016$\bullet$       & .885±.020$\bullet$       & .798±.005$\bullet$                         & .584±.005$\bullet$                         & \multicolumn{1}{l|}{.590±.022$\bullet$} & \textbf{.509±.012} \\
				delicious                                          & .592±.018$\bullet$         & .533±.015$\bullet$       & .618±.017$\bullet$       & .679±.011$\bullet$                         & .452±.007                                  & \multicolumn{1}{l|}{.467±.023$\bullet$} & \textbf{.448±.016} \\ \hline
				\multicolumn{8}{c}{Hamming loss $\downarrow$}   \\
				\hline
				scene                                              & .820±.002$\bullet$         & .820±.003$\bullet$       & .819±.002$\bullet$       & \textbf{.251±.007}       & .814±.000$\bullet$                         & \multicolumn{1}{l|}{.518±.042$\bullet$} & .264±.027                           \\
				yeast                                              & .697±.012$\bullet$         & .697±.013$\bullet$       & .697±.013$\bullet$       & .268±.010$\bullet$                         & .316±.000$\bullet$                         & \multicolumn{1}{l|}{.243±.010}          & \textbf{.235±.008} \\
				eurlex\_dc                                         & .932±.000$\bullet$         & .932±.000$\bullet$       & .118±.006$\bullet$       & .104±.002$\bullet$                         & .890±.039$\bullet$                         & \multicolumn{1}{l|}{.806±.015$\bullet$} & \textbf{.092±.005} \\
				eurlex\_sm                                         & .883±.001$\bullet$         & .883±.001$\bullet$       & .148±.005$\bullet$       & \textbf{.138±.002}        & .825±.027$\bullet$                         & \multicolumn{1}{l|}{.773±.008$\bullet$} & .139±.005                           \\
				corel5k                                            & .886±.007$\bullet$         & .887±.007$\bullet$       & .887±.007$\bullet$       & \textbf{.155±.004}        & .869±.002$\bullet$                         & \multicolumn{1}{l|}{.463±.018$\bullet$} & .229±.068                           \\
				corel16k                                           & .882±.009$\bullet$         & .882±.009$\bullet$       & .882±.009$\bullet$       & \textbf{.177±.011$\circ$} & .862±.001$\bullet$                         & \multicolumn{1}{l|}{.423±.033$\bullet$} & .202±.067                           \\
				bookmark                                           & .917±.001$\bullet$         & .916±.001$\bullet$       & .420±.009$\bullet$       & \textbf{.123±.001$\circ$} & .813±.001$\bullet$                         & \multicolumn{1}{l|}{.409±.014$\bullet$} & .140±.004                           \\
				delicious                                          & .711±.003$\bullet$         & .711±.003$\bullet$       & .711±.003$\bullet$       & .394±.011$\bullet$                         & .459±.002$\bullet$                         & \multicolumn{1}{l|}{.369±.027$\bullet$} & \textbf{.289±.004} \\
				\hline
				\multicolumn{8}{c}{Coverage $\downarrow$}          \\ \hline
				scene                                              & .299±.026$\bullet$         & .256±.017$\bullet$       & .434±.021$\bullet$       & .420±.021$\bullet$                         & \textbf{.230±.006}        & \multicolumn{1}{l|}{.328±.022$\bullet$} & .234±.025                           \\
				yeast                                              & .579±.018$\bullet$         & .649±.020$\bullet$       & .553±.033$\bullet$       & \textbf{.506±.023}        & .742±.027$\bullet$                         & \multicolumn{1}{l|}{.525±.017}          & .525±.021                           \\
				eurlex\_dc                                         & .285±.014$\bullet$         & .269±.015$\bullet$       & .458±.031$\bullet$       & .326±.023$\bullet$                         & .298±.010$\bullet$                         & \multicolumn{1}{l|}{.334±.017$\bullet$} & \textbf{.204±.023} \\
				eurlex\_sm                                         & .416±.010$\bullet$         & .427±.013$\bullet$       & .569±.010$\bullet$       & .509±.013$\bullet$                         & .419±.010$\bullet$                         & \multicolumn{1}{l|}{.519±.008$\bullet$} & \textbf{.365±.014} \\
				corel5k                                            & .473±.034                  & .516±.035$\bullet$       & .529±.028$\bullet$       & .492±.072                                  & \textbf{.429±.008}        & \multicolumn{1}{l|}{.457±.038}          & .445±.048                           \\
				corel16k                                           & .430±.044                  & .488±.027$\bullet$       & .513±.035$\bullet$       & .537±.051$\bullet$                         & .405±.008                                  & \multicolumn{1}{l|}{.407±.033}          & \textbf{.393±.042} \\
				bookmark                                           & .359±.007$\bullet$         & .328±.008$\bullet$       & .475±.019$\bullet$       & .458±.035$\bullet$                         & .280±.004                                  & \multicolumn{1}{l|}{.292±.011$\bullet$} & \textbf{.279±.011} \\
				delicious                                          & .712±.006$\bullet$         & .703±.004$\bullet$       & .726±.009$\bullet$       & .695±.009$\bullet$                         & \textbf{.609±.003$\circ$} & \multicolumn{1}{l|}{.613±.006$\circ$}   & .632±.007                           \\ \hline
				\multicolumn{8}{c}{Average   Precision $\uparrow$}             \\ \hline
				scene                                              & .543±.024$\bullet$         & .600±.017$\bullet$       & .417±.021$\bullet$       & .465±.018$\bullet$                         & .637±.011$\bullet$                         & \multicolumn{1}{l|}{.568±.026$\bullet$} & \textbf{.699±.017} \\
				yeast                                              & .677±.019$\bullet$         & .636±.017$\bullet$       & .688±.017$\bullet$       & .610±.016$\bullet$                         & .459±.032$\bullet$                         & \multicolumn{1}{l|}{.712±.020}          & \textbf{.718±.019} \\
				eurlex\_dc                                         & .412±.018$\bullet$         & .471±.012$\bullet$       & .232±.022$\bullet$       & .373±.015$\bullet$                         & .346±.009$\bullet$                         & \multicolumn{1}{l|}{.250±.031$\bullet$} & \textbf{.549±.025} \\
				eurlex\_sm                                         & .419±.010$\bullet$         & .421±.010$\bullet$       & .273±.006$\bullet$       & .367±.009$\bullet$                         & .402±.005$\bullet$                         & \multicolumn{1}{l|}{.285±.009$\bullet$} & \textbf{.474±.017} \\
				corel5k                                            & .355±.035$\bullet$         & .307±.038$\bullet$       & .297±.023$\bullet$       & .330±.044$\bullet$                         & \textbf{.397±.010}        & \multicolumn{1}{l|}{.371±.028}          & .391±.037                           \\
				corel16k                                           & .405±.050                  & .350±.035$\bullet$       & .325±.022$\bullet$       & .248±.026$\bullet$                         & .424±.006                                  & \multicolumn{1}{l|}{.437±.044}          & \textbf{.449±.049} \\
				bookmark                                           & .383±.007$\bullet$         & .480±.010$\bullet$       & .267±.019$\bullet$       & .329±.016$\bullet$                         & .534±.004$\bullet$                         & \multicolumn{1}{l|}{.506±.014$\bullet$} & \textbf{.584±.013} \\
				delicious                                          & .487±.006$\bullet$         & .511±.004$\bullet$       & .457±.006$\bullet$       & .446±.010$\bullet$                         & \textbf{.580±.002}        & \multicolumn{1}{l|}{.570±.009}          & .572±.005                           \\ \hline
			\end{tabular}
		}
	\end{center}
	\vspace{-0.1in}
\end{table*}

\begin{table*}[t]
	\caption{Experimental results (mean ± std) on training data with biased complementary labels. The best performance of each dataset is presented in \textbf{boldface}, where $\bullet/\circ$ represents whether MLCL is superior/inferior to baselines (with 5\% t-test).}
	\renewcommand\arraystretch{1.2}
	\label{tab_bias}
	\vspace{-0.25in}
	\begin{center}
		\setlength{\tabcolsep}{5mm}{
			\begin{tabular}{llllllll}
				\hline
				\multicolumn{1}{c}{Methods}                       & \multicolumn{1}{c}{ML-KNN} & \multicolumn{1}{c}{LIFT} & \multicolumn{1}{c}{fpml} & \multicolumn{1}{c}{PML-lc}                 & \multicolumn{1}{c}{PML-LRS}                & \multicolumn{1}{c|}{L-UW}               & MLCL					 \\  \hline
				\multicolumn{8}{c}{Ranking loss$\downarrow$} 
				\\ \hline
				scene                                          & \textbf{.086±.015}$\circ$   & .319±.025           & .486±.027$\bullet$                                  & .492±.019$\bullet$                                    & .258±.013$\circ$             & \multicolumn{1}{l|}{.368±.025$\bullet$}                         & .326±.050                                  \\
				yeast                                          & .240±.014$\bullet$                                    & .297±.016$\bullet$                                  & .227±.013$\bullet$                                  & .248±.012$\bullet$                                    & .454±.024$\bullet$                                     & \multicolumn{1}{l|}{.202±.012}  & \textbf{.199±.012}                         \\
				eurlex\_dc                                     & .668±.009$\bullet$                                    & .636±.021$\bullet$                                  & .537±.015$\bullet$                                  & .349±.028$\bullet$                                    & .326±.009              & \multicolumn{1}{l|}{.586±.036$\bullet$}                         & \textbf{.308±.034}                         \\
				eurlex\_sm                                     & .364±.020$\bullet$                                    & .392±.014$\bullet$                                  & .499±.019$\bullet$                                  & .447±.012$\bullet$                                    & .333±.009$\bullet$                                     & \multicolumn{1}{l|}{.641±.015$\bullet$}                         & \textbf{.316±.016}                         \\
				corel5k                                        & \textbf{.324±.038}$\circ$   & .431±.030$\bullet$                                  & .474±.028$\bullet$                                  & .386±.047             & .357±.012              & \multicolumn{1}{l|}{.382±.033}  & .358±.039                                  \\
				corel16k                                       & .413±.063$\bullet$                                    & .431±.041$\bullet$                                  & .454±.033$\bullet$                                  & .471±.068$\bullet$                                    & .375±.015              & \multicolumn{1}{l|}{.373±.029}  & \textbf{.357±.040}                         \\
				bookmark                                       & .567±.007$\bullet$                                    & .449±.042$\bullet$                                  & .552±.018$\bullet$                                  & .491±.016$\bullet$                                    & .244±.003$\bullet$                                     & \multicolumn{1}{l|}{.326±.008$\bullet$}                         & \textbf{.211±.011}                         \\
				delicious                                      & .430±.005$\bullet$                                    & .413±.005$\bullet$                                  & .452±.008$\bullet$                                  & .433±.011$\bullet$                                    & \textbf{.314±.003}$\circ$    & \multicolumn{1}{l|}{.349±.012$\circ$} & .360±.008                                  \\ \hline
				\multicolumn{8}{c}{One Error$\downarrow$}
				\\ \hline
				scene                                          & \textbf{.228±.032}$\circ$   & .669±.043$\bullet$                                  & .803±.038$\bullet$                                  & .720±.018$\bullet$                                    & .613±.017$\bullet$                                     & \multicolumn{1}{l|}{.696±.025$\bullet$}                         & .553±.054                                  \\
				yeast                                          & .330±.032$\bullet$                                    & .280±.025$\bullet$                                  & .254±.028           & .583±.027$\bullet$                                    & .546±.097$\bullet$                                     & \multicolumn{1}{l|}{.256±.025}  & \textbf{.254±.024}                         \\
				eurlex\_dc                                     & .977±.005$\bullet$                                    & .959±.014$\bullet$                                  & .947±.008$\bullet$                                  & .774±.015$\bullet$                                    & .822±.004$\bullet$                                     & \multicolumn{1}{l|}{.822±.038$\bullet$}                         & \textbf{.695±.074}                         \\
				eurlex\_sm                                     & .699±.016$\bullet$                                    & .753±.036$\bullet$                                  & .886±.024$\bullet$                                  & .664±.014             & .737±.011$\bullet$                                     & \multicolumn{1}{l|}{.704±.012$\bullet$}                         & \textbf{.650±.045}                         \\
				corel5k                                        & \textbf{.738±.067}    & .851±.038$\bullet$                                  & .861±.034$\bullet$                                  & .828±.059$\bullet$                                    & .747±.016              & \multicolumn{1}{l|}{.792±.039$\bullet$}                         & .752±.037                                  \\
				corel16k                                       & .780±.061$\bullet$                                    & .827±.049$\bullet$                                  & .837±.025$\bullet$                                  & .952±.021$\bullet$                                    & .730±.000              & \multicolumn{1}{l|}{.731±.053}  & \textbf{.707±.063}                         \\
				bookmark                                       & .906±.007$\bullet$                                    & .804±.037$\bullet$                                  & .925±.008$\bullet$                                  & .792±.004$\bullet$                                    & .576±.003$\bullet$                                     & \multicolumn{1}{l|}{.635±.022$\bullet$}                         & \textbf{.502±.008}                         \\
				delicious                                      & .585±.012$\bullet$                                    & .557±.013$\bullet$                                  & .617±.025$\bullet$                                  & .681±.012$\bullet$                                    & \textbf{.434±.006}$\circ$    & \multicolumn{1}{l|}{.485±.016$\bullet$}                         & .463±.017                                  \\ 
				\hline
				\multicolumn{8}{c}{Hamming loss $\downarrow$}                                                                                                                                                                                                                                                                                                                                                             \\ \hline
				scene                                          & \textbf{.088±.009}$\circ$   & .819±.002$\bullet$                                  & .820±.002$\bullet$                                  & \textbf{.252±.006}    & .814±.000$\bullet$                                     & \multicolumn{1}{l|}{.523±.048$\bullet$}                         & .290±.029                                  \\
				yeast                                          & .697±.012$\bullet$                                    & .697±.013$\bullet$                                  & .697±.013$\bullet$                                  & .268±.010$\bullet$                                    & .316±.000$\bullet$                                     & \multicolumn{1}{l|}{.253±.017$\bullet$}                         & \textbf{.239±.008}                         \\
				eurlex\_dc                                     & .932±.000$\bullet$                                    & .932±.000$\bullet$                                  & .118±.007$\bullet$                                  & \textbf{.104±.002}    & .889±.039$\bullet$                                     & \multicolumn{1}{l|}{.799±.035$\bullet$}                         & .109±.011                                  \\
				eurlex\_sm                                     & .883±.001$\bullet$                                    & .883±.001$\bullet$                                  & .148±.005$\bullet$                                  & .139±.002             & .825±.027$\bullet$                                     & \multicolumn{1}{l|}{.772±.009$\bullet$}                         & \textbf{.138±.007}                         \\
				corel5k                                        & \textbf{.114±.008}$\circ$   & .887±.007$\bullet$                                  & .887±.007$\bullet$                                  & .157±.003$\bullet$                                    & .869±.002$\bullet$                                     & \multicolumn{1}{l|}{.498±.012$\bullet$}                         & .208±.033                                  \\
				corel16k                                       & .882±.009$\bullet$                                    & .882±.009$\bullet$                                  & .882±.009$\bullet$                                  & \textbf{.178±.010}    & .862±.001$\bullet$                                     & \multicolumn{1}{l|}{.481±.028$\bullet$}                         & .207±.086                                  \\
				bookmark                                       & .917±.001$\bullet$                                    & .916±.001$\bullet$                                  & .419±.009$\bullet$                                  & \textbf{.122±.001}$\circ$   & .813±.003$\bullet$                                     & \multicolumn{1}{l|}{.549±.046$\bullet$}                         & .146±.003                                  \\
				delicious                                      & .711±.003$\bullet$                                    & .711±.003$\bullet$                                  & .711±.003$\bullet$                                  & .388±.013$\bullet$                                    & .459±.002$\bullet$                                     & \multicolumn{1}{l|}{.453±.015$\bullet$}                         & \textbf{.304±.005}                         \\
				\hline
				\multicolumn{8}{c}{Coverage$\downarrow$}                                         \\ \hline
				scene                                          & \textbf{.086±.013}$\circ$   & .280±.020           & .420±.023$\bullet$                                  & .420±.016$\bullet$                                    & .229±.011$\circ$             & \multicolumn{1}{l|}{.321±.021$\bullet$}                         & .286±.041                                  \\
				yeast                                          & .551±.017$\bullet$                                    & .638±.028$\bullet$                                  & .533±.012$\bullet$                                  & \textbf{.493±.025}    & .723±.040$\bullet$                                     & \multicolumn{1}{l|}{.500±.018}  & .498±.021                                  \\
				eurlex\_dc                                     & .626±.008$\bullet$                                    & .596±.019$\bullet$                                  & .504±.014$\bullet$                                  & .328±.026$\bullet$                                    & .306±.009$\bullet$                                     & \multicolumn{1}{l|}{.333±.018$\bullet$}                         & \textbf{.274±.030}                         \\
				eurlex\_sm                                     & .432±.018$\bullet$                                    & .456±.014$\bullet$                                  & .579±.015$\bullet$                                  & .520±.015$\bullet$                                    & .418±.009$\bullet$                                     & \multicolumn{1}{l|}{.512±.009$\bullet$}                         & \textbf{.362±.016}                         \\
				corel5k                                        & \textbf{.419±.055}    & .515±.024$\bullet$                                  & .555±.031$\bullet$                                  & .480±.041             & .451±.013              & \multicolumn{1}{l|}{.470±.036}  & .449±.038                                  \\
				corel16k                                       & .498±.052$\bullet$                                    & .521±.038$\bullet$                                  & .542±.035$\bullet$                                  & .533±.066$\bullet$                                    & .454±.018              & \multicolumn{1}{l|}{.468±.030}  & \textbf{.453±.039}                         \\
				bookmark                                       & .565±.006$\bullet$                                    & .455±.039$\bullet$                                  & .553±.017$\bullet$                                  & .492±.014$\bullet$                                    & .265±.003$\bullet$                                     & \multicolumn{1}{l|}{.308±.013$\bullet$}                         & \textbf{.231±.011}                         \\
				delicious                                      & .736±.004$\bullet$                                    & .723±.005$\bullet$                                  & .737±.008$\bullet$                                  & .691±.009             & \textbf{.625±.003}$\circ$    & \multicolumn{1}{l|}{.671±.012$\circ$} & .688±.006                                  \\ \hline
				\multicolumn{8}{c}{Average Precision $\uparrow$}                                                                                                                                                                                                                                                                                                                                                      \\ \hline
				scene                                          & \textbf{.860±.020}$\circ$   & .559±.028$\bullet$                                  & .428±.026$\bullet$                                  & .462±.014$\bullet$                                    & .608±.013              & \multicolumn{1}{l|}{.529±.020$\bullet$}                         & .618±.046                                  \\
				yeast                                          & .670±.023$\bullet$                                    & .634±.016$\bullet$                                  & .691±.022$\bullet$                                  & .614±.015$\bullet$                                    & .500±.026$\bullet$                                     & \multicolumn{1}{l|}{.719±.020}  & \textbf{.726±.018}                         \\
				eurlex\_dc                                     & .145±.005$\bullet$                                    & .166±.016$\bullet$                                  & .201±.009$\bullet$                                  & .371±.020$\bullet$                                    & .357±.005$\bullet$                                     & \multicolumn{1}{l|}{.266±.031$\bullet$}                         & \textbf{.456±.061}                         \\
				eurlex\_sm                                     & .405±.013$\bullet$                                   & .373±.016$\bullet$                                  & .262±.016$\bullet$                                  & .366±.010$\bullet$                                    & .400±.007$\bullet$                                     & \multicolumn{1}{l|}{.282±.011$\bullet$}                         & \textbf{.482±.025}                         \\
				corel5k                                        & \textbf{.409±.040}    & .300±.030$\bullet$                                  & .282±.017$\bullet$                                  & .325±.048$\bullet$                                    & .392±.017              & \multicolumn{1}{l|}{.352±.032}  & .380±.037                                  \\
				corel16k                                       & .355±.054$\bullet$                                    & .318±.033$\bullet$                                  & .301±.024$\bullet$                                  & .240±.030$\bullet$                                    & .393±.054              & \multicolumn{1}{l|}{.384±.036}  & \textbf{.407±.047}                         \\
				bookmark                                       & .219±.004$\bullet$                                    & .320±.037$\bullet$                                  & .212±.007$\bullet$                                  & .320±.004$\bullet$                                    & .544±.003$\bullet$                                     & \multicolumn{1}{l|}{.469±.014$\bullet$}                         & \textbf{.599±.008}                         \\
				delicious                                      & .473±.006$\bullet$                                    & .490±.006$\bullet$                                  & .450±.008$\bullet$                                  & .449±.010$\bullet$                                    & \textbf{.581±.002}$\circ$    & \multicolumn{1}{l|}{.544±.010}  & .544±.009                                  \\ \hline
			\end{tabular}
		}
	\end{center}
	\vspace{-0.1in}
\end{table*}

\begin{table*}[t]
	\caption{Ablation experimental results (mean ± std) on training data with uniform complementary labels. The best performance is in \textbf{boldface}.}
	\renewcommand\arraystretch{1.2}
	\label{tab_ablation}
	\vspace{-0.15in}
	\begin{center}
		\setlength{\tabcolsep}{3.5mm}{
			\begin{tabular}{lcccccccc}
				\hline
				\multicolumn{1}{l|}{\multirow{3}{*}{Methods}} & \multicolumn{4}{c|}{Uniform complementary   labels}                                                    & \multicolumn{4}{c}{Biased complementary   labels}                                 \\ \cline{2-9} 
				\multicolumn{1}{l|}{}                         & scene              & yeast              & eurlex\_dc         & \multicolumn{1}{c|}{corel5k}            & scene              & yeast              & eurlex\_dc         & corel5k            \\ \cline{2-9} 
				\multicolumn{1}{l|}{}                         & \multicolumn{8}{c}{Hamming loss$\downarrow$ }                                                                                                                                                           \\ \hline
				\multicolumn{1}{l|}{MLCL}                     & \textbf{.264±.027} & .235±.008          & \textbf{.092±.005} & \multicolumn{1}{c|}{\textbf{.229±.068}} & \textbf{.290±.029} & .239±.008          & .109±.011          & \textbf{.208±.033} \\
				\multicolumn{1}{l|}{Without $\mathbf C$}                & .290±.039          & .421±.011          & .109±.018          & \multicolumn{1}{c|}{.466±.025}          & .294±.029          & .409±.012          & \textbf{.088±.004} & .444±.031          \\
				\multicolumn{1}{l|}{Without   $\bar L_{mse}$}         & .510±.044          & \textbf{.229±.007} & .509±.043          & \multicolumn{1}{c|}{.461±.053}          & .481±.047          & \textbf{.230±.009} & .512±.046          & .489±.036          \\ \hline
				& \multicolumn{8}{c}{Ranking loss$\downarrow$ }                                                                                                                                                           \\ \hline
				\multicolumn{1}{l|}{MLCL}                     & \textbf{.259±.030} & \textbf{.211±.013} & \textbf{.229±.026} & \multicolumn{1}{c|}{\textbf{.349±.035}} & \textbf{.326±.050} & \textbf{.199±.012} & .308±.034          & \textbf{.358±.039} \\
				\multicolumn{1}{l|}{Without $ \mathbf C $}                & .282±.063          & .419±.018          & .277±.041          & \multicolumn{1}{c|}{.487±.021}          & .348±.046          & .406±.016          & \textbf{.268±.024} & .467±.026          \\
				\multicolumn{1}{l|}{Without   $\bar L_{mse}$}         & .379±.024          & .216±.010          & .303±.028          & \multicolumn{1}{c|}{.362±.030}          & .353±.018          & .204±.011          & .320±.025          & .387±.027          \\ \hline
				& \multicolumn{8}{c}{One error$\downarrow$}                                                                                                                                                              \\ \hline
				\multicolumn{1}{l|}{MLCL}                     & \textbf{.427±.018} & .251±.023          & \textbf{.594±.035} & \multicolumn{1}{c|}{.736±.065}          & \textbf{.553±.054} & \textbf{.254±.024} & .695±.074          & \textbf{.752±.037} \\
				\multicolumn{1}{l|}{Without $ \mathbf C $}                & .474±.047          & .633±.043          & .708±.106          & \multicolumn{1}{c|}{.866±.019}          & .560±.042          & .612±.051          & \textbf{.564±.029} & .855±.027          \\
				\multicolumn{1}{l|}{Without   $\bar L_{mse}$}         & .607±.037          & \textbf{.250±.025} & .740±.048          & \multicolumn{1}{c|}{\textbf{.734±.058}} & .686±.013          & .256±.025          & .753±.044          & .773±.068          \\ \hline
				& \multicolumn{8}{c}{Coverage$\downarrow$}                                                                                                                                                               \\ \hline
				\multicolumn{1}{l|}{MLCL}                     & \textbf{.234±.025} & \textbf{.525±.021} & \textbf{.204±.023} & \multicolumn{1}{c|}{\textbf{.445±.048}}                      & \textbf{.286±.041} & \textbf{.498±.021} & .274±.030          & \textbf{.449±.038} \\
				\multicolumn{1}{l|}{Without $ \mathbf C $}                & .255±.055          & .683±.029          & .247±.035          & \multicolumn{1}{c|}{.565±.032}                               & .306±.039          & .660±.023          & \textbf{.240±.023} & .547±.031          \\
				\multicolumn{1}{l|}{Without   $\bar L_{mse}$}         & .334±.020          & .527±.011          & .249±.024          & \multicolumn{1}{c|}{.451±.035 }                              & .310±.015          & .501±.015          & .265±.022          & .473±.023          \\ \hline
				& \multicolumn{8}{c}{Average precision$\uparrow$}                                                                                                                                                      \\ \hline
				\multicolumn{1}{l|}{MLCL}                     & \textbf{.699±.017} & \textbf{.718±.019} & \textbf{.549±.025} & \multicolumn{1}{c|}{\textbf{.391±.037}} & \textbf{.618±.046} & \textbf{.726±.018} & \textbf{.456±.061} & \textbf{.380±.037} \\
				\multicolumn{1}{l|}{Without $ \mathbf C $}                & .671±.045          & .472±.018          & .469±.085          & \multicolumn{1}{c|}{.274±.014}          & .611±.038          & .489±.015          & .447±.021          & .289±.022          \\
				\multicolumn{1}{l|}{Without   $\bar L_{mse}$}         & .566±.023          & .711±.019          & .426±.040          & \multicolumn{1}{c|}{.389±.041}          & .541±.013          & .717±.020          & .411±.034          & .359±.050          \\ \hline
			\end{tabular}
		}
	\end{center}
\end{table*}

\begin{table*}[t]
	\caption{Parameter sensitivity analysis on uniform complementary-label data, where metric is \emph{average precision}. The best performance is in \textbf{boldface}.}
	\renewcommand\arraystretch{1.2}
	\label{tab_parameter}
	\vspace{-0.15in}
	\begin{center}
		\setlength{\tabcolsep}{4mm}{
			\begin{tabular}{ccccccccc}
				\hline
				$\beta$ & scene     & yeast     & eurlex\_dc & eurlex\_sm & corel5k   & corel16k  & bookmark  & delicious \\ \hline
				0.1     & .678±.017 & .714±.019 & .545±.019  & .451±.025  & .374±.033 & .444±.046 & .565±.007 & .554±.005 \\
				0.3     & .683±.015 & .716±.018 & .549±.021  & .460±.021  & .378±.032 & .447±.047 & .579±.011 & .565±.005 \\
				0.5     & .687±.016 & .718±.018 & .547±.022  & .463±.016  & .385±.031 & .447±.048 & .583±.008 & \textbf{.575±.005} \\
				0.8     & .693±.016 & .718±.018 & .541±.022  & .469±.018  & .387±.037 & .448±.048 & .582±.007 & .572±.006 \\
				1       & \textbf{.699±.017} & \textbf{.718±.019} & \textbf{.549±.025}  & \textbf{.474±.017}  & \textbf{.391±.037} & \textbf{.449±.049} & \textbf{.584±.013} & .572±.005 \\ \hline
			\end{tabular}
		}
	\end{center}
	\vspace{-0.1in}
\end{table*}

\begin{table*}[t]
	\caption{Experimental results (mean ± std) of five criteria.``Fully supervised'' is the linear model training with the fully supervised data (fully supervised MLL). ``CL'' denotes each instance is associated with \textbf{a complementary label} sampled uniformly. ``CL \& RL'' uses the linear model with the loss function Eq.(\ref{eq_13}) to train, where each instance is equipped with \textbf{a complementary label} and \textbf{a relevant label}.}
	\renewcommand\arraystretch{1.2}
	\label{tab_one_label}
	\vspace{-0.15in}
	\begin{center}
		\setlength{\tabcolsep}{3mm}{
			\begin{tabular}{lcccccccc}
				\hline
				Datasets         & scene     & yeast     & eurlex\_dc & eurlex\_sm & corel5k   & corel16k  & bookmark  & delicious \\ \hline
				\multicolumn{9}{c}{Hamming loss$\downarrow$}                                                                                   \\ \hline
				Fully supervised & .120±.013 & .208±.009 & .004±.000  & .033±.001  & .198±.012 & .196±.012 & .098±.004 & .276±.006 \\
				CL               & .264±.027 & .235±.008 & .092±.005  & .139±.005  & .229±.068 & .202±.067 & .140±.004 & .289±.004 \\
				CL \& RL         & .124±.008 & .225±.010 & .005±.001  & .053±.002  & .178±.012 & .172±.010 & .085±.002 & .285±.004 \\ \hline
				\multicolumn{9}{c}{Ranking loss$\downarrow$}                                                                                   \\ \hline
				Fully supervised & .075±.009 & .169±.009 & .003±.001  & .019±.001  & .258±.029 & .222±.029 & .090±.005 & .226±.004 \\
				CL               & .259±.030 & .211±.013 & .229±.026  & .312±.014  & .349±.035 & .289±.042 & .252±.013 & .310±.003 \\
				CL \& RL         & .082±.011 & .191±.011 & .005±.001  & .044±.002  & .268±.031 & .227±.021 & .102±.004 & .267±.004 \\ \hline
				\multicolumn{9}{c}{One Error$\downarrow$}                                                                                      \\ \hline
				Fully supervised & .222±.032 & .223±.023 & .019±.004  & .069±.005  & .627±.038 & .588±.056 & .313±.009 & .340±.012 \\
				CL               & .427±.018 & .251±.023 & .594±.035  & .656±.029  & .736±.065 & .690±.056 & .509±.012 & .448±.016 \\
				CL \& RL         & .229±.033 & .255±.032 & .022±.005  & .098±.007  & .639±.040 & .600±.044 & .324±.007 & .398±.017 \\ \hline
				\multicolumn{9}{c}{Coverage$\downarrow$}                                                                                       \\ \hline
				Fully supervised & .077±.009 & .451±.019 & .004±.000  & .074±.002  & .347±.044 & .315±.024 & .112±.005 & .527±.007 \\
				CL               & .234±.025 & .525±.021 & .204±.023  & .365±.014  & .445±.048 & .393±.042 & .279±.011 & .632±.007 \\
				CL \& RL         & .084±.010 & .474±.021 & .006±.001  & .113±.004  & .363±.048 & .326±.020 & .125±.004 & .564±.006 \\ \hline
				\multicolumn{9}{c}{Average   Precision$\uparrow$}                                                                            \\ \hline
				Fully supervised & .868±.018 & .760±.015 & .988±.003  & .943±.004  & .494±.024 & .530±.038 & .766±.007 & .662±.005 \\
				CL               & .699±.017 & .718±.019 & .549±.025  & .474±.017  & .391±.037 & .449±.049 & .584±.013 & .572±.005 \\
				CL \& RL         & .860±.019 & .734±.018 & .985±.004  & .899±.004  & .485±.028 & .523±.030 & .753±.006 & .618±.005 \\ \hline
			\end{tabular}
		}
	\end{center}
	\vspace{-0.1in}
\end{table*}

\section{Experiments}
\label{sec:experiments}

In this section, we will evaluate the effectiveness of MLCL, where five common MLL criteria, including \textit{ranking loss}, \textit{hamming loss}, \textit{one error}, \textit{coverage} and \textit{average precision}, are employed in this paper. The values of first four criteria are smaller, the performance of approach is better. While the value of \textit{average precision} is greater, the better the performance. The label set of $\bm x$ is predicted by $Y=\{l_k|f^k(\bm x)>0.5, 1\leq k\leq K\}$. All experiments use PyTorch \cite{nips/PaszkeGMLBCKLGA19} and NVIDIA TESLA K80 GPU to implement. The code will be released after this paper has been accepted.

\subsection{Experimental Settings}
\textbf{Datasets.} We use eight widely-used MLL datasets, namely \textit{corel5k}, \textit{corel16k}, \textit{delicious}, \textit{eurlex$\_$dc}, \textit{eurlex$\_$sm}, \textit{yeast}, \textit{bookmarks} and \textit{scene}, to our experiments\footnote{Publicly available at http://mulan.sourceforge.net/datasets.}. Following \cite{aaai/XieH18,aaai/XieH20}, we adopt the same pre-processing to deal with the datasets. More specifically, rare class labels are filtered out for datasets with more than 15 class labels, whose class labels are kept under 15. Accordingly, instances that are relevant with removed class labels are filtered out as well. Detailed characteristics of these datasets are shown in Table \ref{tab_data}.

\textbf{Base models.} The linear model is used as the base model. 

\textbf{Baselines.} Two typical MLL approaches, ML-KNN \cite{pr/ZhangZ07} and LIFT \cite{pami/ZhangW15}, are utilized as baselines, which deal with ML-CLL via regarding all possible labels in the candidate label set as relevant labels for a training instance. Similarly, three recent PML approaches are employed as comparing approaches, including PML-lc \cite{aaai/XieH18}, fpml \cite{icdm/YuCDWLZW18} and PML-LRS \cite{aaai/SunFWLJ19}, which learn from training instances associated with candidate labels. In addition, we employ a multi-class CLL approach, called L-UW \cite{icml/GaoZ21}, as a baseline, which uses BEC loss and \textit{sigmoid} output layer instead of CE loss and \textit{softmax} output layer respectively to make L-UW suit for multi-labeled data.

\subsection{Comparison on Uniform Complementary Labels}
\textbf{Setup.} Weight-decay is set as $1e-4$ and learning rate is selected from $\{1e-1, 1e-2,1e-3\}$ for all data sets. We employ Adam \cite{kingma_adam_2015} optimization method, and set the number of batch-size and epoch as 256 and 200 respectively. L-UW applies the same model and hyper-parameters as ours. Here, we estimate $\mathbf T$ with a linear model. We use Ten-fold cross-validation to evaluate experiments, where training data is associated with complementary labels that are generated by randomly selecting one of possible labels excepting relevant labels (uniform complementary labels), and test data is equipped with the set of relevant labels. The mean metrics value and \textit{standard deviation} (std) will be reported as final experimental results for all approaches.

\textbf{Results.} Table \ref{tab_2} is utilized to report experimental results of various approaches on eight data sets equipped with uniform complementary labels.  $\uparrow/\downarrow$ indicates the larger/smaller the value, the better the performance.

According to reported results in Table \ref{tab_2}, we can observe that results of MLCL are superior or comparable performance against baselines out of different data sets on five criteria. Our approach achieves the best performance in most cases. Specifically, the proposed approach outperforms LIFT on eight datasets across all metrics. This is because our approach is better at tackling the issue that training data is associated with relevant labels and irrelevant labels simultaneously than fully supervised MLL algorithms. Furthermore, experimental results of PML-lc and PML-LRS are inferior to ours in most cases, which demonstrate that PML approaches are indeed inferior to our approach in cases of dense noisy labels. Similarly, based on the results of L-UW shown in Table \ref{tab_2}, we observe that our approach outperforms L-UW on almost all datasets and metrics other than \textit{ranking loss} and \textit{coverage} on the delicious dataset. This reflects that label correlations are important to solve ML-CLL problems, which leads to the proposed approach taking label correlations into account surpasses L-UW that ignores label correlations.

\subsection{Comparison on Biased Complementary Labels}
\textbf{Setup.} To evaluate the effectiveness of our approach in different situations, we utilize training data with biased complementary labels that are generated via the co-occurrence rate of relevant labels. Specifically, we select a complementary label of an instance $\bm x$ from $\mathcal Y\setminus Y$, and the selecting rule follows: the class label with a lower co-occurrence rate has a higher probability to be selected as a complementary label. We adopt training data with biased complementary labels to train the model, while test data is equipped with relevant label sets to evaluate the effectiveness of our approach. For other experimental settings, we apply same settings with Subsection 5.2.

\textbf{Results.} The mean and std of results on test data are shown in Table \ref{tab_bias}. According to results shown in Table \ref{tab_bias}, we can summarize the following impressive observations: (1) MLCL achieves superior or comparable performance to LIFT, fpml, PML-lc, PML-LRS and L-UW on different data sets, which proves that the proposed approach can predict the set of proper labels for unseen instances from complementary labeled data; (2) Although MLCL fails to achieve the best result on the scene dataset, our approach is better than other baselines in the rest of datasets, which indicates that our approach can effectively deal with ML-CLL problems than others. These observations demonstrate that the proposed method can both hold for the situation of data with uniform and biased complementary labels.

\subsection{Additional Experiments}
\textbf{Ablation experiments.} We then explore the effect of different learning components on MLCL performance. Table \ref{tab_ablation} summarizes results of MLCL without the different component, which trains on the  data with uniform complementary labels. In Table \ref{tab_ablation}, without $\mathbf C$ refers to MLCL directly use the estimated initial transition matrix $\mathbf S$ to train, and without $\bar L_{mse}$ indicates that MLCL only utilizes Eq.(\ref{eq_10}) to optimaze. 

From results reported in Table \ref{tab_ablation}, the performance of MLCL surpasses that without different components in most cases, which shows that two components, including using label correlations to correct and an MSE-based regularizer, are beneficial for our approach to improve the performance. Especially, estimating $\mathbf T$ based on label correlations pushes the proposed approach performance forward significantly compared with that without $\mathbf C$ on most cases. Similarly, an MSE-based regularizer brings significant benefits for our approach, which demonstrates that an MSE-based regularizer balances the robustness and convergence rate of BCE loss. These indicate that using label correlations to estimate the transition matrix $\mathbf T$ and an MSE-based regularizer are effective strategies to alleviate ML-CLL problems.

\textbf{Trade-off parameter $\beta$.} Table \ref{tab_parameter} reports the performance of MLCL with varying $\beta$ values that trade-off the complementary loss function $\bar L$ and an MSE-based regularization $\bar L_{mse}$. Here, \textit{average precision} is regarded as the criterion, and the training data is with uniform complementary labels. $\beta$ is selected from the candidate value list $\{0.1, 0.3, 0.5, 0.8, 1\}$. We can observe the best results of most datasets is achieved at $\beta=1$ and the performance drops when $\beta$ takes a smaller value. In general, a relatively large $\beta$ $(\beta\leq 1)$ usually leads to better performance than a small value. Therefore, we set $\beta=1$ for MLCL.

\subsection{Combination of Complementary Labels and Relevant Labels}
\textbf{Setup.} Finally, we demonstrate the effectiveness of combining relevant labeled data and complementary labeled one. The training data is associated with uniform complementary labels and relevant labels simultaneously. More specifically, an instance $\bm x$ is associated with a complementary label $\bar y$ and relevant labels $\tilde{Y}$, where $\bar y$ is uniformly selected and $\tilde{Y}$ is randomly selected from the relevant label set $Y$ of $\bm x$ (i.e., $\tilde{Y} \subseteq Y$). Here, we set $|\tilde{Y}|=1$ that means each instance only associated with a complementary label and a relevant label. The other experimental settings are the same with Subsection 5.2.

\textbf{Results.} We compare three methods: (1) the ``Fully supervised'' method uses the linear model to train with the fully supervised data, which is fully supervised MLL; (2) the ``CL'' method refers to MLCL training with the uniform complementary-label data; (3) the combination (``CL \& RL'') method adopts the linear model with the loss function Eq.(\ref{eq_13}) to train, where the training data is equipped with the combination of complementary labels and relevant labels. Table \ref{tab_one_label} reports the experimental results on five criteria. We can see that the performance of ``CL\& RL'' method is much superior to ``CL'' method on all datasets over \textit{hamming loss}, \textit{ranking loss}, \textit{one error}, \textit{coverage} and \textit{average precision}, such as ``CL\& RL'' method outperforms ``CL'' method by a large margin over \textit{average precision} (\textbf{+0.436} on eurlex\_dc and \textbf{+0.425} on eurlex\_sm). This demonstrates that the ML-CLL is easily applied to fully supervised MLL scenarios, MLL with missing labels \cite{tcyb/FengHSA22,apin/WangLL19} or other MLL scenarios. Moreover, ``CL \& RL'' method achieves comparable performance to ``Fully supervised'' method, which illustrates that ML-CLL can get excellent results just via increasing a few additional information. This is useful for application in the real world, because ML-CLL can obtain good performance through less expensive labeled data.

\section{Conclusion}
\label{sec:conclusion}
In this paper, we theoretically analyze the reason causing why the estimated transition matrix in multi-class CLL is distorted in ML-CLL. To alleviate the pain in directly calculating the transition matrix from complex label correlations under multi-labeled data is unknown, we propose a two-step method to estimate the transition matrix $\mathbf T$ in ML-CLL, which adopts label correlations to correct an initial transition matrix. Furthermore, we theoretically show that the proposed approach is classifier-consistent. Additionally, due to MSE loss achieving a prominent robust, an MSE-based regularizer is introduced to alleviate the tendency of the fast convergent BCE loss overfitting to noises. Finally, we show that our proposed ML-CLL can be easily combined with relevant labels and the proposed method can achieve a comparable performance to fully supervised MLL through a few additional information.


\ifCLASSOPTIONcaptionsoff
\newpage
\fi



%
%
%

\bibliographystyle{IEEEtran}
\bibliography{example_paper}

%




\appendices
\onecolumn

\section{The Proof of Theorem \ref{theo:matrix}}
\label{apx:theorem_1}
\textbf{Theorem \ref{theo:matrix}.} \textit{Given an instance $\bm x$, suppose $Y$ is the relevant label set and the label $l_j$ is the complementary label which is randomly selected. Then the following equality holds: 
	\begin{align*}
		p(\bar y^j=1|\bm x)=\sum_{C\in \mathcal{Y}', l_j\notin C} p(\bar y^j=1| Y=C)p(Y=C|\bm x)
		\ge \sum_{k=1, k\neq j}^{K} p(\bar y^j=1|y^k=1)p(y^k=1|\bm x).
\end{align*}}
\begin{proof}
	Firstly, we should introduce addition rule of probability: $p(AB)=p(A)+p(B)-p(A \cup B)$, so we have $p(AB)\geq p(A)+p(B)$. We start to prove the above inequlity. According to the assumption: $p(\bar y|Y)=p(\bar y|Y, \bm x)$, we have
	\begin{align*}
		p(\bar y^j=1|\bm x)&=\sum_{C\in \mathcal{Y}', l_j\notin C} p(\bar y^j=1| Y=C)p(Y=C|\bm x)
		\\
		&=\sum_{C\in \mathcal{Y}', l_j\notin C} p(\bar y^j=1| Y=C,\bm x)p(Y=C|\bm x)
		\\
		&=\sum_{C\in \mathcal{Y}', l_j\notin C} p(\bar y^j=1, Y=C|\bm x)
		\\
		&=\sum_{C\in \mathcal{Y}', l_j\notin C} p(Y=C|\bar y^j=1, \bm x)p(\bar y^j=1| \bm x). 
	\end{align*}
	According to addition rule of probability, so we have
	\begin{align*}
		p(\bar y^j=1|\bm x)&\geq \sum_{C\in \mathcal{Y}', l_j\notin C} \left[\sum_{k=1, k\neq j, l_k\in C}^{K}p(y^k=1|\bar y^j=1, \bm x) + \sum_{k=1, l_k\notin C}^{K}p(y^k=0|\bar y^j=1, \bm x) \right] p(\bar y^j=1| \bm x)
		\\
		&\geq \sum_{C\in \mathcal{Y}', l_j\notin C} \sum_{k=1, k\neq j, l_k\in C}^{K}p(y^k=1|\bar y^j=1, \bm x) p(\bar y^j=1| \bm x) \;\;\;\;\;\; \text{$\because \sum_{k=1, l_k\notin C}^{K}p(y^k=0|\bar y^j=1, \bm x)\geq 0$}
		\\
		&=\sum_{C\in \mathcal{Y}', l_j\notin C} \sum_{k=1, k\neq j, l_k\in C}^{K}p(\bar y^j=1 |y^k=1, \bm x)p(y^k=1| \bm x)
		\\
		&=\sum_{C\in \mathcal{Y}', l_j\notin C} \sum_{k=1, k\neq j}^{K}p(\bar y^j=1|y^k=1, \bm x)p(y^k=1| \bm x) \;\;\;\;\;\; \text{$\because p(y^k=1|\bm x)=0$ if $\l_k\notin Y$ }
		\\
		&=\sum_{k=1, k\neq j}^{K}\sum_{C\in \mathcal{Y}', l_j\notin C} p(\bar y^j=1|y^k=1, \bm x)p(y^k=1| \bm x)
		\\
		&=\sum_{k=1, k\neq j}^{K} (2^{K-1}-1)p(\bar y^j=1|y^k=1, \bm x)p(y^k=1| \bm x)
		\\
		&\geq \sum_{k=1, k\neq j}^{K} p(\bar y^j=1|y^k=1, \bm x)p(y^k=1| \bm x)
		\\
		&=\sum_{k=1, k\neq j}^{K} p(\bar y^j=1|y^k=1)p(y^k=1| \bm x).
	\end{align*}
\end{proof}

\section{The Proof of Theorem \ref{theo_T}}
\label{apx:estimate}

\textbf{Theorem \ref{theo_T}.} 
\textit{Under a MLL scenario: suppose the labels $l_{z_1}, l_{z_2}\in\mathcal Y$ ($z_1, z_2 \in [K], z_1\neq z_2$) are dependent, and the labels belonging to $\mathcal Y\setminus\{l_{z_1},l_{z_2}\}$ are mutually exclusive. For any $\bm x\in \mathcal{X}$, its label set $Y \subseteq \{l_{z_1},l_{z_2}\}$ and $Y \neq \emptyset$. Let the label $l_j$ ($j\in[K], j\neq z_1,z_2$) be the complementary label of $\bm x$. $\mathbf T_{z_1j}$ and $\mathbf T_{z_2j}$ calculated from label correlations satisfy
	\begin{align*}
		\mathbf T_{z_1j}=\frac{p(\bar y^j=1|\bm x)}{p(y^{z_2}=1|\bar y^j=1,y^{z_1}=1, \bm x)p(y^{z_1}=1|\bm x)}, 
		\\
		\mathbf T_{z_2j}=\frac{p(\bar y^j=1|\bm x)}{p(y^{z_1}=1|\bar y^j=1,y^{z_2}=1, \bm x)p(y^{z_2}=1|\bm x)},
	\end{align*}
	where $[K]$ denotes the integer set $\{1,2,\dots,K\}$. The difference of $\mathbf{T}$ and $\mathbf{Q}$ on the complementary label $l_j$ is
	\begin{align*}
		\ell_j \geq 2(\frac{1}{\xi^2}-1)p(\bar y^j=1|\bm x),
	\end{align*}
	where $\xi=\max \{ p(y^{z_2}=1|\bar y^j=1,y^{z_1}=1, \bm x),p(y^{z_1}=1|\bar y^j=1,y^{z_2}=1, \bm x) \}$.}
\begin{proof}
	We start calculating the difference $\ell_j$ from estimating the transition probabilities $\mathbf{T}_{z_1 j}$ and $\mathbf{T}_{z_1 j}$. According to Definition 2 and the description of Theorem \ref{theo_T}, we have
	\begin{align*}
		p(\bar y^j=1|\bm x)&=\sum_{k=1,k\neq j,z_1,z_2}^K p(\bar y^j=1|y^k=1,\bm x)p(y^k=1|\bm x)+p(\bar y^j=1|y^{z_1}=1,y^{z_2}=1,\bm x)p(y^{z_1}=1,y^{z_2}=1|\bm x)\\
		&+p(\bar y^j=1|y^{z_1}=1,y^{z_2}=0,\bm x)p(y^{z_1}=1,y^{z_2}=0|\bm x) + p(\bar y^j=1|y^{z_1}=0,y^{z_2}=1,\bm x)p(y^{z_1}=0,y^{z_2}=1|\bm x)\\
		&+p(\bar y^j=1|y^{z_1}=0,y^{z_2}=0,\bm x)p(y^{z_1}=0,y^{z_2}=0|\bm x)
		\\
		&=\sum_{k=1,k\neq j,z_1,z_2}^K p(\bar y^j=1|y^k=1,\bm x)p(y^k=1|\bm x) +p(y^{z_2}=1|\bar y^j=1,y^{z_1}=1,\bm x)p(\bar y^j=1|y^{z_1}=1,\bm x)p(y^{z_1}=1|\bm x)\\
		&+p(y^{z_1}=1|\bar y^j=1,y^{z_2}=0,\bm x)p(\bar y^j=1|y^{z_2}=0,\bm x)p(y^{z_2}=0|\bm x)\\
		&+p(y^{z_2}=1|\bar y^j=1,y^{z_1}=0,\bm x)p(\bar y^j=1|y^{z_1}=0,\bm x)p(y^{z_1}=0|\bm x)\\
		&+p(y^{z_2}=0|\bar y^j=1,y^{z_1}=0,\bm x)p(\bar y^j=1|y^{z_1}=0,\bm x)p(y^{z_1}=0|\bm x).
	\end{align*}
	
	Based on the assumption of that $\bar y$ and $\bm x$ are conditionally independent given $Y$, then we can have
	\begin{align*}
		p(\bar y^j=1|\bm x)&=\sum_{k=1,k\neq j,z_1,z_2}^K p(\bar y^j=1|y^k=1)p(y^k=1|\bm x) +p(y^{z_2}=1|\bar y^j=1,y^{z_1}=1,\bm x)p(\bar y^j=1|y^{z_1}=1)p(y^{z_1}=1|\bm x)\\
		&+p(y^{z_1}=1|\bar y^j=1,y^{z_2}=0,\bm x)p(\bar y^j=1|y^{z_2}=0)p(y^{z_2}=0|\bm x)\\
		&+p(y^{z_2}=1|\bar y^j=1,y^{z_1}=0,\bm x)p(\bar y^j=1|y^{z_1}=0)p(y^{z_1}=0|\bm x)\\
		&+p(y^{z_2}=0|\bar y^j=1,y^{z_1}=0,\bm x)p(\bar y^j=1|y^{z_1}=0)p(y^{z_1}=0|\bm x).
	\end{align*}
	
	Since $p(\bar y^j=1|y^{z_1}=0)$ and $p(\bar y^j=1|y^{z_2}=0)$ do not hold according to the definition of the transition matrix, and then we can obtain
	\begin{align*}
		p(\bar y^j=1|\bm x)&=\sum_{k=1,k\neq j,z_1,z_2}^K p(\bar y^j=1|y^k=1)p(y^k=1|\bm x) +p(y^{z_2}=1|\bar y^j=1,y^{z_1}=1,\bm x)p(\bar y^j=1|y^{z_1}=1)p(y^{z_1}=1|\bm x)
		\\
		&=p(y^{z_2}=1|\bar y^j=1,y^{z_1}=1,\bm x)p(\bar y^j=1|y^{z_1}=1)p(y^{z_1}=1|\bm x)  \;\;\;\;\;\; \text{$\because p(y^k=1|\bm x)=0$ if $l_k\notin Y$}
	\end{align*}
	\begin{align*}
		\Rightarrow \mathbf T_{z_1j}=p(\bar y^j=1|y^{z_1}=1)=\frac{p(\bar y^j=1|\bm x)}{p(y^{z_2}=1|\bar y^j=1,y^{z_1}=1, \bm x)p(y^{z_1}=1| \bm x)}.
	\end{align*}
	
	Similarly, we can get 
	\begin{align*}
		\mathbf T_{z_2j}=p(\bar y^j=1|y^{z_2}=1)=\frac{p(\bar y^j=1|\bm x)}{p(y^{z_1}=1|\bar y^j=1,y^{z_2}=1, \bm x)p(y^{z_2}=1| \bm x)}.
	\end{align*}
	
	Next, we calculate the difference $\ell_j$. The rest elements of $\mathbf{T}_{\cdot j}$ are same as that estimated by multi-class CLL. According the definition of $\ell_j$, we have
	\begin{align*}
		\ell_j &= \sum_{k=1}^K|\mathbf T_{kj}-\mathbf Q_{kj}|
		\\
		&=\left|\mathbf T_{z_1j}+\mathbf T_{z_2j}-2p(\bar y^j=1|\bm x)\right|
		\\
		&=\left|\frac{p(\bar y^j=1|\bm x)}{p(y^{z_2}=1|\bar y^j=1,y^{z_1}=1, \bm x)p(y^{z_1}=1| \bm x)} + \frac{p(\bar y^j=1|\bm x)}{p(y^{z_1}=1|\bar y^j=1,y^{z_2}=1, \bm x)p(y^{z_2}=1| \bm x)} - 2p(\bar y^j=1|\bm x) \right|
		\\
		&\geq\left|2(\frac{1}{\xi^2}-1)p(\bar y^j=1|\bm x)\right|  
		\\
		&=2(\frac{1}{\xi^2}-1)p(\bar y^j=1|\bm x). \;\;\;\; \text{$\because \frac{1}{\xi^2}\geq 1$}
	\end{align*}
	
	Because $0\leq p(y^{z_1}=1| \bm x) \leq p(y^{z_1}=1|\bar y^j=1,y^{z_2}=1, \bm x)\leq 1$ and $0\leq p(y^{z_2}=1| \bm x) \leq p(y^{z_2}=1|\bar y^j=1,y^{z_1}=1, \bm x)\leq 1$, $\xi$ is defined as $\xi = \mathrm{max}\{p(y^{z_2}=1|\bar y^j=1,y^{z_1}=1, \bm x),p(y^{z_1}=1|\bar y^j=1,y^{z_2}=1, \bm x)\}$, the above inequation holds.
\end{proof}

\section{The Proof of Corollary \ref{coro}}
\label{apx:cor}
\textbf{Corollary \ref{coro}.} \textit{Under a MLL scenario: there are $m$ ($m\geq 2$) labels $l_{z_1}, l_{z_2},\dots,l_{z_m} \in \mathcal Y$ $(z_1,\dots, z_m\in [K])$ that are dependent, while the labels belong to $\mathcal Y\setminus\{l_{z_1},l_{z_2}, \dots, l_{z_m}\}$ are mutually exclusive. For any $\bm x \in\mathcal{X}$, its relevant set $Y \subseteq \{l_{z_1}, l_{z_2}\dots, l_{z_m}\}$ and $Y \neq \emptyset$. Suppose the label $l_j$ is the complementary label of $\bm x$. The difference $\ell_j$ between $\mathbf{T}$ and $\mathbf{Q}$ has
	\begin{align*}
		\ell_j\geq m(\frac{1}{\xi^{m}}-1)p(\bar y^j=1|\bm x),
	\end{align*}
	where $\xi = \mathrm{max}\{p(y^{z_m}=1|\bar y^j=1,y^{z_1}=1,\dots,y^{z_{m-1}}=1, \bm x), p(y^{z_{m-1}}=1|\bar y^j=1,y^{z_1}=1,\dots,y^{z_{m-2}}=1, y^{z_{m}}=1, \bm x),\dots, p(y^{z_1}=1|\bar y^j=1,y^{z_2}=1,\dots,y^{z_{m}}=1, \bm x)\}$ $(\xi\in(0,1])$.}
\begin{proof}
	Here, we apply induction to get the difference as $m$ increases. We start by computing the difference in the case of $m=3$. Suppose class labels $l_{z_1},l_{z_2},l_{z_3}\in\mathcal Y$ are dependent, while the rest of labels in the label space are mutually exclusive. $\bm x$ is associated with $Y\subseteq\{l_{z_1},l_{z_2},l_{z_3}\}$ and $Y\neq \emptyset$. Then we calculate transition probabilities in $\mathbf T$ from label correlations according to Theorem \ref{theo_T} as:
	\begin{align*}
		p(\bar y^j=1|\bm x)&= \sum_{k=1,k\neq j,z_1,z_2,z_3}^K p(\bar y^j=1|y^k=1,\bm x)p(y^k=1|\bm x)+p(\bar y^j=1,y^{z_1}=1,y^{z_2}=1,y^{z_3}=1|\bm x) 
		\\
		&=p(\bar y^j=1,y^{z_1}=1,y^{z_2}=1,y^{z_3}=1|\bm x)
		\\
		&=p(y^{z_3}=1|\bar y^j=1, y^{z_1}=1, y^{z_2}=1, \bm x)p(y^{z_2}=1|\bar y^j=1, y^{z_1}=1,\bm x)p(\bar y^j=1| y^{z_1}=1,\bm x)p(y^{z_1}=1|\bm x)
		\\
		&=p(y^{z_3}=1|\bar y^j=1, y^{z_1}=1, y^{z_2}=1, \bm x)p(y^{z_2}=1|\bar y^j=1, y^{z_1}=1,\bm x)p(\bar y^j=1| y^{z_1}=1)p(y^{z_1}=1|\bm x)
	\end{align*}
	\begin{align*}
		\Rightarrow \mathbf T_{z_1j}=p(\bar y^j=1| y^{z_1}=1)= \frac{p(\bar y^j=1|\bm x)}{p(y^{z_3}=1|\bar y^j=1, y^{z_1}=1, y^{z_2}=1, \bm x)p(y^{z_2}=1|\bar y^j=1, y^{z_1}=1,\bm x)p(y^{z_1}=1|\bm x)}.
	\end{align*}
	
	$\mathbf T_{z_2j}$ and $\mathbf T_{z_3j}$ use the same way to estimate. Due to $0\leq p(y^{z_1}=1| \bm x) \leq p(y^{z_1}=1|\bar y^j=1,y^{z_2}=1, \bm x)\leq  p(y^{z_1}=1|\bar y^j=1,y^{z_2}=1, y^{z_3}=1, \bm x) \leq 1$, let $\xi=\mathrm{max}\{p(y^{z_3}=1|\bar y^j=1, y^{z_1}=1, y^{z_2}=1, \bm x), p(y^{z_2}=1|\bar y^j=1, y^{z_1}=1, y^{z_3}=1, \bm x), p(y^{z_1}=1|\bar y^j=1, y^{z_2}=1, y^{z_3}=1, \bm x)\}$, we can obtain
	\begin{align*}
		\mathbf T_{z_1j}= p(\bar y^j=1| y^{z_1}=1)\geq \frac{1}{\xi^3}p(\bar y^j=1|\bm x).
	\end{align*}
	
	Similarly, we can compute $\mathbf T_{z_2j}, \mathbf T_{z_3j}\geq \frac{1}{\xi^3}p(\bar y^j=1|\bm x)$. Then the difference $\ell_j$ is 
	\begin{align*}
		\ell_j &= \sum_{k=1}^K \left|\mathbf T_{kj}-\mathbf Q_{kj}\right|
		\\
		&=\left|\mathbf T_{z_1j}+T_{z_2j}+\mathbf T_{z_3j}-3p(\bar y^j=1|\bm x)\right|
		\\
		&\geq 3(\frac{1}{\xi^3}-1)p(\bar y^j=1|\bm x).
	\end{align*}
	
	Similarly, for any $m$ $(0<m<K)$, suppose class labels $l_{z_1},l_{z_2},\dots,l_{z_m}\in\mathcal Y$ are strongly dependent, while the rest of labels in the label space are mutually exclusive. $\bm x$ is associated with $Y\subseteq\{l_{z_1},l_{z_2},l_{z_3}\}$ and $Y\neq \emptyset$. Then we calculate transition probabilities from label correlations:
	\begin{align*}
		p(\bar{y}^j=1|\bm x)&=\sum_{k=1,k\neq j,z_1,\dots,z_m}^K p(\bar{y}^{j}=1|y^{k}=1, \bm x) p(y^{k}=1|\bm x)+p(\bar{y}^{j}=1, y^{z_{1}}=1, \dots, y^{z_{m}}=1|\bm x)
		\\
		&=p(\bar{y}^{j}=1, y^{z_{1}}=1, \dots, y^{z_{s}}=1|\bm x)
		\\
		&=p(y^{z_{2}}=1|\bar{y}^{j}=1, y^{z_{1}}=1, \bm{x})p(\bar{y}^{j}=1|y^{z_{1}}=1) p(y^{z_{1}}=1|\bm{x}) \Pi_{i=3}^{m} p(y^{z_i}=1|\bar{y}^{j}=1, y^{z_{1}}=1, \dots, y^{z_{i-1}}=1, \bm{x})  
	\end{align*}
	\begin{align*}
		\Rightarrow \mathbf T_{z_1j} = p(\bar y^j=1| y^{z_1}=1)=\frac{p(\bar y^j=1|\bm x)}{p(y^{z_{2}}=1|\bar{y}^{j}=1, y^{z_{1}}=1, \bm{x})p(y^{z_1}=1|\bm x)\Pi_{i=3}^{m} p(y^{z_i}=1|\bar{y}^{j}=1, y^{z_{1}}=1, \dots, y^{z_{i-1}}=1, \bm{x})}.
	\end{align*}
	
	As discussed above, $\mathbf T_{z_1j}\geq \frac{1}{\xi^{m}}p(\bar y^j=1|\bm x)$ since $\xi = \mathrm{max}\{p(y^{z_m}=1|\bar y^j=1,y^{z_1}=1,\dots,y^{z_{m-1}}=1, \bm x), p(y^{z_{m-1}}=1|\bar y^j=1,y^{z_1}=1,\dots,y^{z_{m-2}}=1, y^{z_{m}}=1, \bm x),\dots, p(y^{z_1}=1|\bar y^j=1,y^{z_2}=1,\dots,y^{z_{m}}=1, \bm x)\}$ $(\xi\in(0,1])$. By the same calculation way, we can obtain $\mathbf T_{z_2j}, \dots, \mathbf T_{z_mj} \geq \frac{1}{\xi^{m}}p(\bar y^j=1|\bm x)$. Based on induction, we can summarize the difference $\ell_j =\sum_{k=1}^K |\mathbf T_{kj}-\mathbf Q_{kj}| \geq m(\frac{1}{\xi^{m}}-1)p(\bar y^j=1|\bm x)$. 
\end{proof}

\section{The Proof of Theorem 6}
\label{apx:consist}
\textbf{Theorem 6.} \textit{With Assumption \ref{assumption}, suppose the transition matrix $\mathbf T$ is invertible, then the ML-CLL optimal classifier $\bm f_{CL}^*$ converges to the MLL optimal classifier $\bm f^*$, i.e., $\bm f_{CL}^*=\bm f^*$.}
\begin{proof}
	We prove $\bm f^*$ is also the optimal classifier for ML-CLL via substituting $\bm f^*$ into the ML-CLL risk:
	\begin{align*}
		R_{\bar L}(\bm f^*)&=\mathbb E_{p(\bm x, \bar y)}[\bar L(\bm f^*(\bm x),\bm{\bar y})]
		\\
		&=\int \sum_{\bar y\in\mathcal{Y}} \bar L(\bm f^*(\bm x),\bm{\bar y}) p(\bm x, \bar y) d\bm x
		\\
		&=\int \sum_{\bar y\in\mathcal{Y}} L(\mathbf T^T\bm f^*(\bm x),\bm{\bar y}) \sum_{Y\in\mathcal{Y}}p(Y|\bar y,\bm x)p(\bar y, \bm x) d\bm x 
		\\
		& = \int \sum_{\bar y\in\mathcal{Y}} \sum_{Y\in\mathcal{Y}} L(\mathbf T^T\bm f^*(\bm x),\bm{\bar y}) p(\bar y| Y,\bm x)p(Y, \bm x) d\bm x
		\\
		&=\int \sum_{Y\in\mathcal{Y}} L(\mathbf T^T\bm f^*(\bm x),\bm{\bar y})p(Y, \bm x) d\bm x
		\\
		&=\int \sum_{Y\in\mathcal{Y}} L(\mathbf T^T\bm f^*(\bm x), T^T\bm y)p(Y, \bm x) d\bm x
		\\
		&=R(\mathbf T^T\bm f^*)
	\end{align*}
	
	According to the proof of \cite{eccv/yu_learning_2018}, $\bm f_{CL}^*=\mathbf T^T\bm f^*$. So we find the optimal $\bm f^*$ ensuring $\bm f_{CL}^*=\bm f^*$ when the transition matrix $\mathbf T$ is invertible and Assumption \ref{assumption} is satisfied.
\end{proof}

\end{document}